# Twenty Years of Personality Computing: Threats, Challenges and Future Directions

FABIO CELLI, Research and Development, Maggioli SpA, Italy

ALEKSANDAR KARTELJ, Faculty of Mathematics, University of Belgrade, Serbia

MILJAN ĐORĐEVIĆ, School of Electrical Engineering, University of Belgrade, Serbia

DERWIN SUHARTONO, School of Computer Science, Bina Nusantara University, Indonesia

VLADIMIR FILIPOVIĆ, Faculty of Mathematics, University of Belgrade, Serbia

VELJKO MILUTINOVIĆ, School of Electrical Engineering, University of Belgrade, Serbia

GEORGIOS SPATHOULAS, Dept. of Information Security, Norwegian University of Science and Technology, Norway

ALESSANDRO VINCIARELLI, School of Computing Science, University of Glasgow, UK

MICHAL KOSINSKI, Graduate School of Business, Stanford University, USA

BRUNO LEPRI, Center for Augmented Intelligence, Fondazione Bruno Kessler, Italy

Personality Computing is a field at the intersection of Personality Psychology and Computer Science. Started in 2005, research in the field utilizes computational methods to understand and predict human personality traits. The expansion of the field has been very rapid and, by analyzing digital footprints (text, images, social media, etc.), it helped to develop systems that recognize and even replicate human personality. While offering promising applications in talent recruiting, marketing and healthcare, the ethical implications of Personality Computing are significant. Concerns include data privacy, algorithmic bias, and the potential for manipulation by personality-aware Artificial Intelligence. This paper provides an overview of the field, explores key methodologies, discusses the challenges and threats, and outlines potential future directions for responsible development and deployment of Personality Computing technologies.

Authors' Contact Information: Fabio Celli, Research and Development, Maggioli SpA, Santarcangelo di Romagna, Italy, fabio.celli@maggioli.it; Aleksandar Kartelj, Faculty of Mathematics, University of Belgrade, Belgrade, Serbia; Miljan Đorđević, School of Electrical Engineering, University of Belgrade, Belgrade, Serbia; Derwin Suhartono, School of Computer Science, Bina Nusantara University, Jakarta, Indonesia; Vladimir Filipović, Faculty of Mathematics, University of Belgrade, Belgrade, Serbia; Veljko Milutinović, School of Electrical Engineering, University of Belgrade, Belgrade, Serbia; Georgios Spathoulas, Dept. of Information Security, Norwegian University of Science and Technology, Gjøvik, Norway; Alessandro Vinciarelli, School of Computing Science, University of Glasgow, Glasgow, UK; Michal Kosinski, Graduate School of Business, Stanford University, Stanford, CA, USA; Bruno Lepri, Center for Augmented Intelligence, Fondazione Bruno Kessler, Trento, Italy.

## 1 Introduction

Personality Computing is a research field started in 2005 that uses computational methods to understand and predict human personality. There are many digital footprints conveying information that computational models can decode and use in various ways to model behavior and interact with humans. These include text [171, 173, 174], images [86, 244, 245], videos and audio tracks [130, 190, 218, 265], social media posts and likes [153, 155, 242], credit card transactions [99, 259], mobile phone communication logs [52, 64, 192, 249], experience sampling apps [177], and physiological data [76]. The basis for training Personality Computing models comes from Personality Psychology [230], primarily using questionnaires to assess and quantify personality types and scores [37, 61]. The availability of data that combine digital footprints and scores from personality assessments allowed the development of systems that recognize human personality exploiting sophisticated computational techniques, including Natural Language Processing (NLP), Computer Vision (CV), Automatic Speech Recognition (ASR), Deep Learning (DL) and Social Network Analysis (SNA). Systems for Personality Computing have been applied in different domains, like personalized marketing and digital mass persuasion [239, 240], recommendation systems [67], mental health detection [294], prediction of entrepreneurial outcomes [18, 151] and Human-Computer Interaction tasks [24, 25].

Crucially, these algorithms had a powerful persuasion effect on voters in different elections [290], raising serious concerns about the ethical use of Personality Computing. In particular, the data protection regulations (e.g., the European Union General Data Protection Regulation - GDPR[1]) and the legislation on Artificial Intelligence (e.g., the European Union AI Act[2]) often limit Personality Computing. There are restrictions on the types of data that can be collected, rules prohibiting uses that could be discriminatory or harmful, and laws that require systems to be transparent about how personality inferences are made, and to provide explanations for automated decisions involving personality. Recent advancements in Artificial Intelligence (AI) have led to Large Language Models (LLMs), which are able to detect and replicate human personality during an interaction and role-playing tasks [120, 124, 125, 133, 134, 214, 269, 270], potentially reaching a high level of intimacy between AI systems and humans. This might yield to scenarios where AI systems can manipulate humans to engineer large scale social functions, such as voting and spending, with a disruptive impact on human societies [115]. This is why a survey of this field is essential to identify challenges and explore ways to mitigate the negative social impacts of Personality Computing. This paper is structured as follows: first we introduce the definition of personality traits and the notion of personality states. In Section 2 we take an overview of the field, its history, benchmarks and the State-of-the-Art (SotA). In Section 3 we discuss methods and techniques to build Personality Computing systems with different types of data sources, from text to multimedia and mobile phone traces. Finally, we draw our conclusions in Section 4, where we discuss threats and challenges in Personality Computing, and outline possible future directions to mitigate negative effects for societies. We believe that this paper could be a valuable resource for future generations of researchers in personality computing and more broadly of researchers interested in building AI agents able to interact and collaborate with humans.

---

[1]https://gdpr-info.eu/
[2]https://artificialintelligenceact.eu/

## 2 Background, History and Resources

### 2.1 What is Personality?

Personality is a complex system of individual characteristics that strongly affects a person's thoughts, emotions, and behavior in any given situation. It's visible in both digital and real-world interactions, and it significantly shapes one's self-image, expectations, values, and attitudes [136]. Personality Psychology is a branch of psychology that aims to explain the mental processes of personality [230] and provide frameworks for the assessment of personality [136]. Research in Personality Psychology began one century ago and produced different methods for assessing personality types (see Section 2.2). The availability of widely accepted personality assessment methods, both self-reported and observer-rated, facilitated the collection of data annotated with personality scores and labels, serving as a gold standard in Personality Computing research. Another important and more recent concept is the notion of "psychological states", sometimes referred to as "states of personality" [92, 93]. In psychology, traits and states are often seen as opposite ends of a continuum. Unlike personality traits, which are relatively stable and enduring characteristics, states are transient and influenced by the immediate context [58, 89, 93]. A person may typically be introverted (trait), but in a highly social and exciting environment where they feel at ease, they might exhibit extraverted behavior (state) without changing personality. States of personality provides a more nuanced understanding of human behavior by acknowledging the dynamic interplay between stable dispositions and situational influences. Understanding psychological states is important in all subfields of psychology, from social to organizational psychology, for example to analyze employee behavior in specific work contexts. Research also demonstrated that the nature of social interactions can significantly influence the expression of personality states in individuals. In particular, adaptation effects were observed for agreeable and emotionally stable people. Individuals with these states tend to evoke similar states in others during interactions [11]. Another essential aspect of personality is how it is perceived. The Brunswik lens model provides a framework for understanding the process of interpersonal perception. This model defines the relationship between observable cues (distal cues) and the ultimate judgment or perception (proximal percept). It emphasizes the probabilistic nature of this relationship, acknowledging that cues are not perfectly reliable indicators. In personality research, the lens model helps explain how individuals infer personality traits from observable behaviors, and acknowledges that these inferences are based on probabilistic cues and that accuracy can vary. The model is used to show how subjective cognitive and emotional states can form from objective quantifiable behaviors [274].

### 2.2 Personality models

There are several models and assessments to measure personality. In Personality Computing, the most used personality assessments are questionnaires, that can map a set of items to specific personality traits and facets. For example, the item "I am reserved" is correlated with introversion. Of course questionnaires have biases, for example the "social desirability" bias [156]: People taking personality tests often try to present themselves in a certain way, which can skew the results. While no single technique completely eliminates bias from personality tests, there are methods to statistically control it, like Social Desirability Scales [63], randomized response and Lie scoring [129].

There are two main models used in Personality Computing: the Five-Factor Model of personality (FFM), evolved in parallel with the Big Five model [10, 104, 211], and the MBTI (Myers-Briggs Type Indicator) model [193]. There are also other models, such as DISC [206], Hogan [119] and the Short Dark Triad (SD3) [138], but they are used in specific contexts and are not common in Personality Computing.

The Big Five and FFM models independently arrived to very similar conclusions about personality traits [199, 209] and quantify personality into the following five traits:

(1) Openness to experience – intellectual and insightful versus shallow and unimaginative;
(2) Conscientiousness – self-disciplined and organized versus inefficient and careless;
(3) Extraversion – sociable, assertive and playful versus aloof, reserved and shy;
(4) Agreeableness – friendly and cooperative versus antagonistic and faultfinding;
(5) Neuroticism – confident and calm versus anxious and insecure.

The Big Five, sometimes referred to as OCEAN, is based on a lexical approach to personality measurement [103, 180] and has been shown to hold across different ages, genders, and cultural lines [135, 179]. Additional research has shown that different tests, languages, and analysis methods do not alter the validity of the model [68, 135, 178]. Such extensive research has led many psychologists to adopt the Big Five, and it is the most widely accepted model of personality in academia [210]. The Big Five has been replicated in a variety of different languages and cultures, such as Chinese [260] and Indian [168]. However, some researchers suggest that the Openness trait is unsupported in some Asian cultures like the Chinese, the Japanese and the Philippine, and that a different sixth factor is sometimes identified [35, 50]. In its six-factor extension, OCEAN becomes HEXACO (Honesty-Humility, Emotionality, eXtraversion, Agreeableness, Conscientiousness and Openness), but this model is rarely used in the literature [15].

A very controversial personality test, well known outside of academia, is the MBTI. It is based on Carl Jung's personality typology [139] and classifies profiles into 16 categories based on the combination of 4 binary traits:

(1) extroversion/introversion – introverts reflect before they act, while extroverts act before they think;
(2) sensing/intuition – sensing trust concrete information, while intuitive trust abstract and theoretical information;
(3) thinking/feeling – feeling decide based on emotions, while thinking involve logic and reason in their decisions;
(4) judging/perceiving – judging prefer structure in their lives, while perceiving like change.

Despite Personality Psychology literature consistently shows that there is no validity to it [38], MBTI is sometimes adopted by computer scientists for the availability of data annotated with this model, and for the fact that its categorical output, as opposed to scalar scores, facilitates the use of classification algorithms.

## 2.3 History of Personality Computing

For many years researchers in Personality Psychology used small samples with few tens or hundreds subjects in their experimental settings. Between the 1950s and the 1970s part of the research was focused on personality measurement [83], the validity of personality tests [105] and its real-life applications [198]. Moreover there was also a rising awareness that personality and language are connected [106, 108]. The many different personality assessment tests developed in more than 60 years (see Section 2.4), allowed researchers in the field of Personality Computing to collect large datasets annotated with personality scores and types (see Section 2.5 for details) for training computational models. The demonstration that personality can be computed from language dates back to 1999 [213], but the first automatic classification of personality from text appeared in 2005 [254]. In our opinion, this marked the beginning of Personality Computing as a new field of research at the intersection of psychology and computer science. At the beginning, in the field of NLP, personality scores were often turned into categories with mean or median split, and personality recognition was addressed as a multi-class classification task. Since then, Personality Computing has expanded from NLP, becoming a field of research that bridges the gap between Personality Psychology, Affective Computing and AI.

Systems in Personality Computing use computational techniques to analyze digital footprints, from text messages to multimedia and other sources of data, to solve three key problems [265]:

(1) **Automatic Personality Recognition**, the task of predicting personality based on the persons' digital records like texts, pictures, audio, multimedia, mobile phone traces, social media likes, etc.

(2) **Automatic Personality Perception**, whose aim is to computationally understand and evaluate how personality is perceived from humans and machines.

(3) **Automatic Personality Synthesis**, the task to create and evaluate personalities for virtual characters like avatars and AI agents.

Figure 1, taken from [265], depicts the relationship between the Brunswik Lens and the three main problems addressed in Personality Computing.

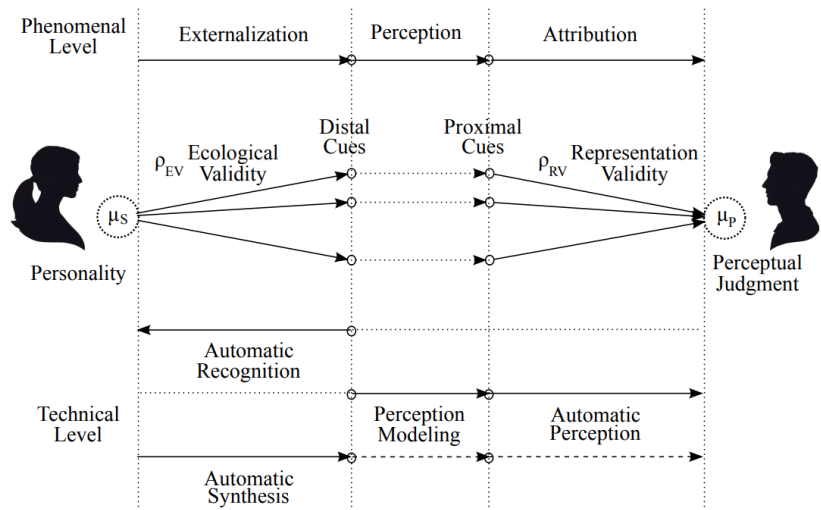

Fig. 1. Relationship between the Brunswik Lens and tasks in Personality Computing. The picture is taken from [265].

The Stimulus (S) is the true, underlying personality of an individual, while Ecological Validity (EV) refers to how well the observable behavior in the environment or digital footprints (distal cues) actually reflect the true personality. For example, how well someone's clothing choices or profile picture selection reflect their personality. Proximal Cues are the processed and interpreted cues that an observer perceives. These are influenced by the observer's own biases, experiences, and perceptual filters. For this reason Representation Validity (RV) is different from the Ecological Validity, and measures how accurately the observer's internal representation of the cues matches the actual proximal cues. Perceptual Judgment (P) is the observer's final judgment or perception of personality, based on the processed information. The model highlights challenges for Personality Computing. One is the ecological validity of the data used. Digital footprints may not always accurately reflect an individual's true personality and personality recognition can measure this association. Personality Synthesis can measure how well a system can generate ecologically valid distal cues, while Personality Perception should measure the relationship between distal cues, proximal cues, and perceptual judgments. A Personality Computing system or AI agent that performs these three tasks with high accuracy, is potentially able to manipulate humans, understanding, simulating and adapting to our personality. For this reason it is extremely important that researchers working in Personality Computing are aware of the complexities of human perception to design and develop more secure and ethical personality computing systems.

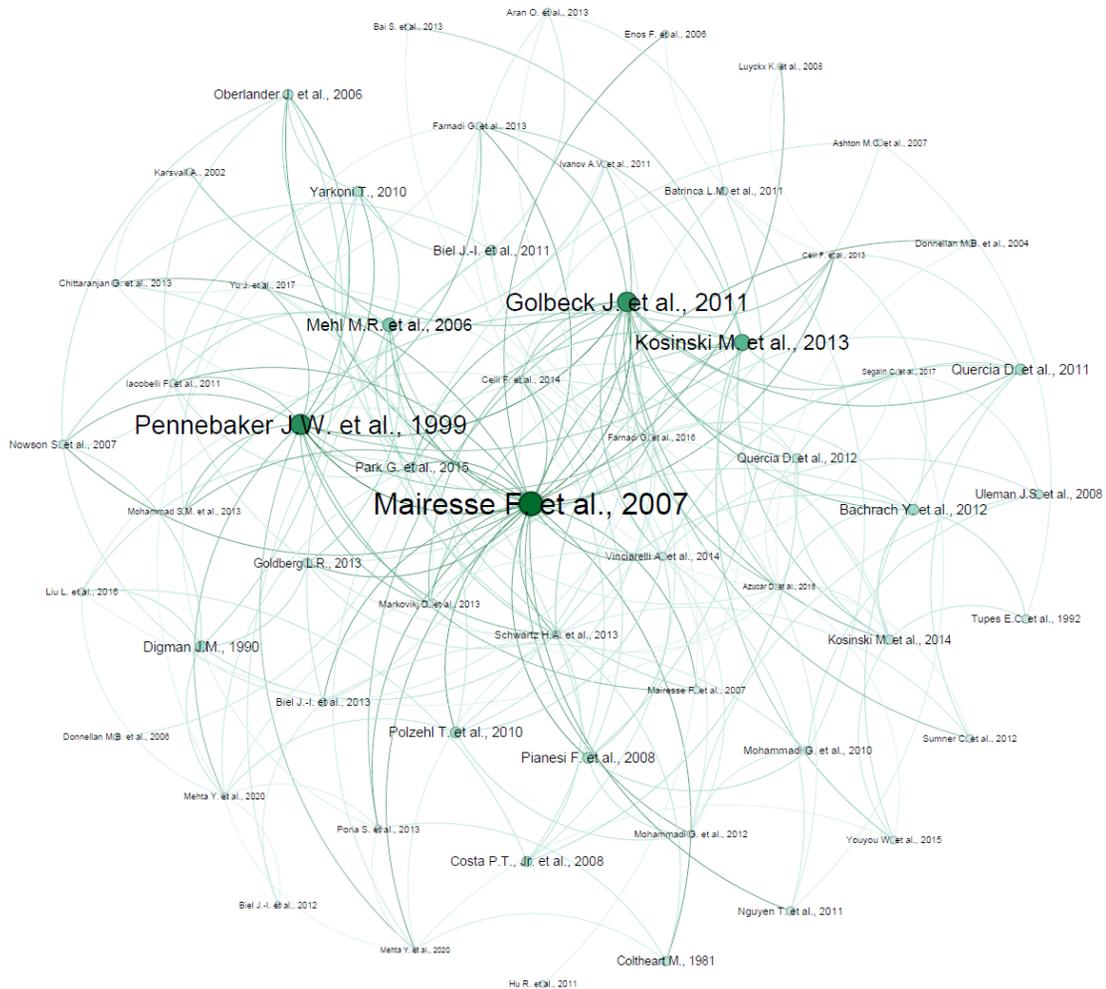

Fig. 2. Scopus citation network of the works cited in this paper with in-degree above 5.

The field of Personality Computing has seen a rapid expansion in these 20 years, with a very rapid growth around Personality Recognition in the first 10 years and a growing interest towards the other Personality Computing tasks in more recent years. In order to reconstruct the roots of Personality Computing, we built a citation network of the works cited in this paper with in-degree above 5. The graph, reported in Figure 2, is not meant to be exhaustive or representative of all the branches in the field, but it highlights the seminal papers that bridged the gap between psychology and computer science. As already mentioned, the first influential paper was done by Pennebaker and King [213], who created the first dataset (Essays) and the first resource (Linguistic Inquiry and Word Count - LIWC). Another influential paper was done by Mehl et al. [182], who started a different line of research that observes how personality is shown and interpreted during normal, daily activities. However, Mairesse et al. [174] bridged the gap between psychology and computer science. They used text and audio corpora, extracted features with psycholinguistic resources and classified personality with machine learning algorithms. Although they did not obtain high classification performances (below 62% of accuracy), they showed that linguistic resources were very useful to extract interpretable features for psycholinguistic models that could be effectively investigated with computational methods on a large scale. After this work emerged two branches. One branch focused on the recognition of personality from multimodal information (mainly the fusion of visual and acoustic features) [217]. The second one started applying personality recognition to social media data, in particular Facebook and Twitter [101, 102, 222]. Along this line, Kosinski et al. [155] showed that Facebook likes were extremely informative about users' personality. Moreover, they were the first to raise concerns about how Personality Recognition threatens users' privacy. The fastest-growing task of Personality Computing in the 2010s was Personality Recognition from text, and two distinct methods emerged: the top-down approach, that consists in the exploitation of psycholinguistic resources to extract features from text, and the bottom-up approach, that consists in the extraction of features with unsupervised techniques such as bag-of-words. Although this latter approach was in use since the first studies on Personality Computing and author profiling in blogs [203, 286], the most influential works using a bottom-up approach are the study of Bachrach et al. [19], who extracted correlations between personality and Facebook metadata, and the one of Schwartz et al. [242], who analyzed the vocabulary of specific personality types from 700 million words of Facebook users. These works were possible thanks to MyPersonality[3], the largest Facebook dataset annotated with personality scores (see Section 2.5). In 2012 Mohammadi and Vinciarelli [190] made significant advances in the field of Personality Recognition by using audio prosodic features. The Shared tasks in 2013 and 2014 [44, 46] provided one of the first comparison arena between approaches, datasets and algorithms in the task of Personality Recognition from text and multimedia. Results showed that the mix of top-down and bottom-up approaches perform best; ensemble learning is very good to detect weak signals from features and spontaneous data from social media is more informative than elicited data. Some influential papers of 2013 were participants to the shared task [82, 176]. In 2014 the first complete survey about personality computing was written by Vinciarelli and Mohammadi [265]: this survey defined tasks and challenges and had a great impact for the diffusion of knowledge in the field. Another influential paper is the one done by Park et al. [208] who used again MyPersonality dataset to confirm that language-based assessments can constitute valid personality measures. In the same year, Youyou et al. [288] found that computer-based models outperform humans in assessing personality from social media data. A work that has high in-degree centrality in the network, shown in Figure 2, is the one by Farnadi et al. [80], who performed a comparative analysis of SotA methods for computational Personality Recognition on different social media datasets, including text from Facebook and Twitter and multimedia from Youtube. They reported SotA results for classification tasks, that predict personality traits as binary classes, as well as for regression tasks, that treat the personality traits as scales. In the following years Personality Recognition exploded in China, with influential papers by [279, 280] that predicted personality scores from small Chinese text datasets of the social media Sina Weibo. They use Label Distribution Learning (i.e., a machine learning approach for minimizing the distance between the model output and the ground-truth label distribution [266]) and Deep Learning methods, obtaining results that cannot be compared directly to previous literature. Nevertheless, they set a new trend with the use of Deep Learning for the recognition of personality [293] from different digital footprints, including profile pictures [243]. In 2018 Azucar et al. [17] made a systematic meta-analysis to determine the predictive power of different digital footprints, including pictures, social media activity and demographics. They found that accuracy of predictions is consistent across Big 5 traits, and that accuracy improves when analyses include demographics and multiple types of digital footprints processed together. In the following years techniques for Personality Computing from social media were replicated in other languages, such as Bahasa [204]. In order to provide a good performance of Personality Recognition in such different languages, some adaptation and optimization approaches were experimented [3]. these include semi-supervised learning [2], Deep Learning with BERT and ULMFiT [114], the development of specific resources like IndoBERT [148], and pre-trained language models with ensemble majority voting [194]. In sum, the past research in Personality Computing tended to focus on the following aspects: the type of personality traits to predict (Big Five, MBTI); the type of data and, in the case of text, the language; the type of features (bottom-up/top-down approach); the type of algorithms and the the type of task (classification, regression).

---

[3]https://psycnet.apa.org/fulltext/2015-39598-005.html

Table 1. Most commonly used personality questionnaires

| Questionnaire | References | Item count |
|---|---|---|
| TIPI | [136] | 10 |
| Mini-IPIP | [70] | 20 |
| IPIP-IPC-32 | [175] | 32 |
| Hogan Personality Inventory | [119] | 42 |
| BFI-10 | [22] | 44 |
| NEO-FFI | [60] | 60 |
| MBTI | [193] | 94 |
| IPIP-FFM | [276] | 100 |
| EPQ-R | [30] | 100 |
| HEXACO-PI-R | [160] | 100 |
| IPIP-NEO-120 | [137] | 120 |
| BFAS | [94] | 40 |
| Woodworth Psychoneurotic Inventory | [207] | 116 |
| 16PF | [41] | 185 |
| California Psychological Inventory | [107] | 194 |
| NEO-PI-R | [60, 61] | 240 |

### 2.4 Personality Questionnaires

Questionnaires are the key instrument to produce gold standard labeled data for Personality Computing. Measuring personality traits is usually achieved by either explicit, self-reported assessments or implicit, observer-rating assessments. Explicit methods traditionally measure personality using a series of items or questions selected from an inventory. The first ever text-based questionnaire, The Woodworth Psychoneurotic Inventory [207], was developed in the US during World War I for the purpose of screening recruits for shell shock risks. Personality tests evolved through time for better modeling personality, and from the Coopersmith self-esteem inventory [56], 16PF [41] and EPQ-R, that contains 48 questions [30], are also utilized for more specialized purposes. The most commonly used text-based questionnaires are presented in Table 1. As previously said, there are two main personality assessment methods: self-assessment and observer-assessment. The former requires that the subject takes the test, while the latter can be compiled by another person, the observer, and it is also possible to have two or more external observers completing the test in order to have more robust assessments [55]. It is worth noting that the observer assessment method may require monitoring sensitive behavioral data which can expose private conversations or other confidential data, with potential privacy issues. Personality tests can be very short (10 items) to very long (340 items). On the Internet, however, relying on long text-based questionnaires is often impractical [121]. For this reason, researchers proposed short questionnaires like the Ten Item Personality Inventory (TIPI) [136], a visual test [236] and a couple of videogame-based personality assessments [113, 224, 282], but their utilization is yet to reach the levels of classical approaches. Online environments have proven to be quite rich when it comes to asserting user's characteristics. [47, 81, 208, 222]. Some of the most used social networks in Personality Computing are Facebook, Twitter, Instagram, Linkedin, Youtube, and the chinese WeChat, Sina Weibo and RenRen [20]. The characteristics of the social networks environment and the impossibility to control the context when users take the tests online may result in invalid assessments or churn. This led to the use of very short personality tests designed for the web, but there are works that show how the use of very short measures of personality may substantially increase both the Type 1 and Type 2 error rates [62].

Table 2. Main Benchmarks and Resources for Personality Computing, by type, language, available data points (Persons) and dimensions (Features).

| Benchmarks | Type | Public | Language | Persons | Features | Reference |
|---|---|---|---|---|---|---|
| Essays | text | no | eng | ≈ 2,500 | 1 | [213] |
| EAR | audio | no | eng | 96 | 1 | [174] |
| Speaker Personality Corpus | audio | no | eng | 11 | ≈ 100 | [173] |
| PersIA | audio | no | eng | 22 | 6,552 | [128] |
| MyPersonality corpus | mixed | yes | mixed | ≈ 6,000,000 | ≈ 50 | [153] |
| Mission Survival Corpus | mixed | no | ita | 48 | 22 | [217] |
| IS2012 speaker trait challenge | audio | no | eng | 322 | 6,125 | [241] |
| WCPR2013 | text | yes | eng | 250 | 18 | [46] |
| WCPR2014 | mixed | yes | eng | ≈ 460 | ≈ 40 | [44] |
| PAN-AP-2015 | text | yes | multi | 294 | 9 | [229] |
| MBTItweets | text | yes | eng | 1500 | 2 | [219] |
| Big5Mbti | text | yes | multi | 1,800 | 1,078 | [43] |
| PANDORA talks | text | yes | eng | ≈ 10,000 | 1 | [98] |
| WASSA | text | yes | eng | ≈ 1,000 | ≈ 20 | [23] |
| Personality Simulation 1 | tabular | yes | - | 400 | ≈ 100 | [270] |
| Resources | Type | Public | Language | Entries | Features | |
| LIWC | text | no | multi | 12,000 words | 88 | [212] |
| MRC | text | yes | eng | 150,837 words | 26 | [54] |
| OpenSMILE | audio | yes | - | - | 15 | [77] |
| Computational Aesthetics | image | no | - | - | 16 | [245] |

## 2.5 Benchmarks and Resources

Personality Computing requires large amounts of data annotated with personality types, but collecting data of this kind is difficult and costly. For this reason, there are few very large datasets and many small ones. A summary of benchmarks and resources for Personality Computing used in literature is reported in Table 2.

The first large corpus annotated with personality that became a benchmark is Essays [213], a collection of self descriptions by students who completed the Big 5 tests. It contains only elicited text and features are not pre-extracted. Given a great interest in web blogs and social media data, the MyPersonality project [154] started collecting personality data together with digital footprints from Facebook between 2007 and 2015. With 6 millions users collected, this is the largest corpus annotated with personality ever produced, and contains different types of assessments, network structures, Facebook likes, profile pictures, demographic information and other types of digital footprints. MyPersonality has been used in many works [19, 40, 79, 82, 112, 153, 155, 223] and had a great impact on research in the field. The Mission Survival Corpus [217] is the first multimodal (audio and video tracks) dataset annotated with personality, the first audio corpus is instead the Speaker Personality Corpus (SPC) [173], followed by the Speaker Trait Challenge dataset [241]. The Workshop on Computational Personality Recognition 2013 (WCPR2013) [46] and 2014 (WCPR2014) [44] were the first shared tasks that released publicly available corpora for personality recognition. WCPR2013 released a corpus with social media text data and network structure of a small and anonymized part of MyPersonality, while WCPR2014 released two datasets: one with mobile data and one with vlogs [31, 32] containing transcriptions and features extracted from video. The PAN Author Profiling shared task in 2015 (PAN-AP-2015) [229] contained personality annotations for 294 Twitter users and their posts. Personality traits were self-assessed with the BFI-10[22]. PersIA [128] is an audio dataset with simulated tourist call center conversations annotated with the personality of the caller. After the introduction of a very fast search method for automatically extracting textual information annotated with MBTI personality tests from Twitter [219] and from the PersonalityCafé forum [235], a very interesting collected dataset was Big5Mbti [43], the first to contain text data of Twitter users annotated both with Big 5 and MBTI labels. Another dataset annotated with both labels is PANDORA talks, a large corpus of text taken from Reddit [98]. The Popularity of MBTI personality types, especially outside of academia, is clear in the fact that The MBTI-type dataset is freely available on Kaggle for competions. Interestingly, a very recent trend is to generate data using Large Language Models (LLMs), that are able to produce synthetic data with the characteristics of general personality types, like depressive, obsessive, paranoid or narcissistic, but not specific Big Five personality scores [195]. Since 2022, the WASSA competition includes a shared task on Personality Recognition from text [23], among other tasks like empathy and emotion recognition. Finally, the Personality Simulation 1 dataset, designed for Personality Synthesis, has been recently released in the APA repository [270]. Among the resources used for Personality Computing with top-down methods, one of the most important is Linguistic Inquiry and Word Count (LIWC). LIWC was originally developed as a dictionary mapping words to psycholinguistic dimensions, such as the use of first-person, negations, pronouns, and so on. LIWC is now a software [36, 212] that provides APIs in more than one language. MRC (Medical Research Council) is a psycholinguistic database [54, 273] that contains a dictionary of 98538 words. MRC feature types are counts of nouns, adjectives, verbs, adverbs, prepositions, conjunctions, pronouns, interjections, past participle, and others. There are also resources for top-down feature extraction from multimedia. OpenSMILE [77] is a common software used for the extraction of prosodic and technical features from audio and speech. Computational Aesthetics [245] is an extraction tool that maps images into easily interpretable visual aspects, like color diversity, number of details and rule of thirds. The same tools for the extraction of features from images can be used in the extraction from videos, adding the time feature. The bottom-up approach does not rely on proper resources, but rather on code libraries and algorithms, like n-grams, bag-of-words or features extracted with Convolutional Neural Networks (CNN). More recent approaches rely on LLMs. LLMs, trained on massive text and code datasets, learn intricate patterns in human language, including personality. Combining top-down and bottom up approaches, LLMs can address Personality Synthesis tasks as well as Personality Recognition even without any example (zero-shot prompting).

## 2.6 State of the Art

It is very difficult to report the SotA in Personality Computing because scholars and even shared tasks adopted different evaluation metrics and different experimental settings. Table 3 summarizes the SotA in all tasks of Personality Computing. Personality Recognition task is simple: given a set of personality assessments and related digital footprints, this task predicts personality scores or classes from the digital footprints, like text or multimedia. The first shared tasks, like the InterSpeech Speaker Trait Challenge 2012, the WCPR2013 and 2014, and the PAN Author Profiling used different metrics for the evaluation averaged over the five traits, including averaged F1-score, averaged accuracy and averaged Root Mean Squared Error (RMSE) [98]. The WASSA competiton adopted averaged correlation coefficients (rho), that is rapidly becoming the standard to evaluate Personality Recognition. In 2024 the highest average correlation was 0.3, over a baseline of 0.133, but a constant improvement is observed from previous editions [97]. Recent findings report that GPT-3.5 and GPT-4, despite not being specifically trained for personality prediction, can infer the Big Five personality traits from users' Facebook status updates with an average correlation of 0.29 with zero-shot prompting. Although there are differences between traits with Conscientiousness having lower performance, the averaged result is comparable to that of sophisticated machine learning models explicitly designed for personality inference [214]. A replication study on the PAN-AP-2015 task with the T5 LLM [225] and zero-shot prompting obtained an averaged F1-measure of 0.711 [271], that is in line with most of the systems participating to the competition, but lower with respect to the best one.

Table 3. SotA in Personality Computing. Metrics include F1-measure (F1) and Accuracy (Acc) for classification evaluations, Root Mean Squared Error (RMSE), Cronbach's $\alpha$ (Alpha) and correlation coefficients (Rho) for regression evaluations. Results are averaged over all the personality traits. *Effect sizes must be interpreted with caution; the human perception and computer studies were performed using different methods (Schmidt–Hunter and multilevel modeling, respectively), data sets, and analytical approaches that may not be fully comparable.

| Personality recognition | Data and target | F1 | Acc | RMSE | Rho |
|---|---|---|---|---|---|
| IS2012 speaker trait challenge | audio, Big5 | | 0.683 | | |
| WCPR2013 | text, Big5 | 0.720 | 0.563 | | |
| WCPR2014 | mixed, Big5 | 0.670 | 0.620 | 0.760 | |
| PAN-AP-2015 | text, Big5 | 0.711 | 0.840 | | |
| Big5mbti | text, MBTI | | 0.649 | | |
| WASSA2022 | text, Big5 | | | | 0.230 |
| WASSA2023 | text, Big5 | | | | 0.252 |
| WASSA2024 | text, Big5 | | | | 0.300 |
| PANDORA talks | text, Big5+MBTI | 0.660 | | | 0.292 |
| Peters & Matz 2024 | text, Big5 | | | | 0.290 |
| Personality synthesis | Data and target | F1 | Acc | Alpha | Rho |
| Klinkert et al. 2024 | GPT-4, Big5 | | 0.740 | | |
| Wang et al. 2024 | GPT-4, Big5 | | | 0.980 | |
| Personality perception | Data and target | F1 | Acc | Alpha | Rho |
| Youyou et al. 2015 | social media, Big5 | | | | 0.560 |
| Mohammadi & Vinciarelli 2015 | audio, Big5 | | 0.720 | | |
| Hinds et al. 2024 | mixed, Big5 | | | | *0.300 |

Personality Synthesis task is more complex: a set of ground truth personality scores are used to create role-play setting prompts, that a LLM will execute. The evaluation is done by asking to the LLM to complete a self-assessment assuming each role-played character and measuring the consistency of the assessment results. Although there are different metrics, such as averaged accuracy and averaged Cronbach's alpha, research in Personality Synthesis demonstrated that LLMs are able to generate realistic and engaging personalities in role-play settings and gaming [152], with incredibly high internal consistency, and with a more structured factor organization compared to the human counterparts they were based on [270]. Personality Perception is the inference of personality that others attribute to an individual based on observable behavior. The Personality Perception task is the most complex one, and consists in the comparison between human and machine judgments on personality from digital footprints. In this task, personality is defined as self-other agreement (i.e., effect sizes) of different human raters, and results are usually measured with Pearson correlations or using the Kendall's Coefficient of Concordance (also called Kendall's W) [149]. Sometimes, the results are also compared against the scores provided by self-assessments, to understand the difference between self-perception and others' perception. As mentioned above, Youyou et al. [288] revealed that computer-generated personality judgments correlated more strongly (correlation coefficient = 0.56) with individuals' self-perceptions than judgments made by humans (correlation coefficient = 0.49). Furthermore, computer models demonstrated greater consistency among different assessments (higher inter-rater reliability) and stronger predictive validity [288]. Mohammadi and Vinciarelli [190] investigated the use of prosodic features (e.g., pitch, rhythm, intonation) in the perception of speakers' personality. They find that machines can perceive personality with an accuracy of 72% with respect to humans from audio files. They report that Extraversion and Conscientiousness tend to be predicted better, in line with the observations of Personality Psychology. Hinds and Joinson [118] made an extensive meta analysis of more than 30 studies on Personality Perception. Their research investigates the convergent validity of self-reported personality with both human perception and Personality Computing systems. Meta-analyses of human perception studies found moderate convergent validity across the Big Five traits (rho = 0.38). Similarly, a meta-analysis of computer prediction studies demonstrated moderate convergent validity (rho = 0.30). Effect sizes of this study must be interpreted with caution as the human and computer perception experimental settings used different methods, data sets, and approaches that may not be fully comparable. Despite this, the meta-analysis identified that data sources (e.g., social media, videos, smartphone data) have a significative impact on the variance explained. In other words, the communication channels affect the way personality is perceived. Under this perspective, the medium has an impact on the message [181].

## 3  Personality Computing Systems

Personality Computing systems have different components, including pre-processing, feature extraction, and classification/generation. Typically, systems that employ machine learning should be designed and trained in all the distinct parts, systems based on Deep Learning automate the pre-processing and feature extraction part, while LLM-based systems have a complete automation, but are less transparent. In this section, we present an overview of the main Personality Computing systems, divided into text-based, image-based, audio-based, video-based, hybrid systems (based on other sources of data such as social media metadata, mobile phone data, etc.) and LLM-based systems.

### 3.1  Text-based systems

In order to provide an extensive overview of Personality Computing methods and systems, we reviewed 45 relevant papers published between 2006 and 2023. Table 4 summarizes this information, categorizing papers by year, algorithms, data type, and features. We grouped similar algorithms (e.g., AdaBoost [91], XGBoost [49], and Gradient Boosting into the 'Boosting' category) and applied similar groupings to data types and features. Our analysis revealed that, among the systems for Personality Recognition from text, Support Vector Machines (SVMs) [59] were the most frequently used algorithms (20 papers) before the introduction of Neural Networks (12 papers). Surprisingly, Naive Bayes are well represented (10 papers). Notably, the use of Neural Networks has significantly increased in recent years. Regarding data sources, social network data is the most frequent source, appearing in 27 papers. Essays were also a frequently used benchmark, in 15 studies. While weblogs were prominent in earlier years, their usage has declined since 2011, likely mirroring their decreasing popularity as an online platform. Psycho-linguistic features were the most common homogeneous feature category (16 papers), although a slightly higher number of papers (20) utilized a broader range of general word-based features. In terms of research activity, a relatively consistent trend is observed, with a slight surge between 2011 and 2013 and a renewed increase after 2019.

Table 4. Text-based systems

| Published | Reference | Bayesian multinomial regression | Boosting | Conditional random fields | Decision tree | Gaussian process | Lazy learners | Logistic regression | Linear Discriminant Analysis | Memory-based learning | Naive Bayes | Nearest neighbors | Neural networks | Random forest | Receptiviti API | Ridge regression | Rule-based learners | Sequential minimal optimization | Support vector machine | Stochastic Gradient descent | Blogs and weblogs | Electronically Activated Recorder | Emails | Essays | MyPersonality | Social networks | Metadata and activity | Non-lexical synthetic | N-grams | Other lexical (word-based) | Part-Of-Speech | Psycho-linguistic (LIWC, MRC....) | Semantic |
|---|---|---|---|---|---|---|---|---|---|---|---|---|---|---|---|---|---|---|---|---|---|---|---|---|---|---|---|---|---|---|---|---|---|
| | | Algorithms | | | | | | | | | | | | | | | | | | | Data type | | | | | | Features | | | | | | |
| 2006 | [203] | | | | | | | | | | ✓ | | | | | | | | | | | | | | | | | | | | | ✓ | |
| 2006 | [173] | | | | ✓ | | | | | | | | | | | | | | | | | ✓ | ✓ | | | | | | | | | | ✓ |
| 2007 | [75] | | ✓ | | ✓ | ✓ | | | | | | | | ✓ | | | | | ✓ | | ✓ | ✓ | | | | | | | | | ✓ | ✓ | |
| 2008 | [170] | | | | | | | | | ✓ | | | | | | | | | | | | ✓ | | | | | | | | | ✓ | ✓ | |
| 2009 | [13] | ✓ | | | | | | | | | | | | | | | | | | | ✓ | ✓ | | | | | | | | | ✓ | ✓ | |
| 2011 | [186] | | | | | | | | ✓ | | | | | | | | | | | | ✓ | | | | | | | | | | | ✓ | |
| 2011 | [222] | | | | ✓ | | | | | | | | | | | | | | | | | | | | | ✓ | ✓ | | | | | | |
| 2011 | [102] | | | | | ✓ | | | | | | | | | | | | | | | | | | | | ✓ | | | | | | | |
| 2011 | [196] | | | | | | | | | | ✓ | | | | | | | | ✓ | | ✓ | | | | | | | | | | | ✓ | |
| 2011 | [127] | | | | | | | | | | | | | | | | | | ✓ | | | | | ✓ | | | | | ✓ | | | ✓ | |
| 2012 | [253] | | | | ✓ | | | | | | ✓ | | | ✓ | | | | | ✓ | | | | | ✓ | | | | | | | | ✓ | |
| 2012 | [150] | | | | | | | | | | | | | | | | | | ✓ | | | | | | ✓ | | | | | | ✓ | ✓ | |
| 2013 | [48] | | | | | | | ✓ | | | | | ✓ | | | | | ✓ | | | | | | ✓ | | | | | | ✓ | | ✓ | |
| 2013 | [221] | | | | | | | | | | ✓ | ✓ | | | | | | | | | | | | ✓ | | | | | | ✓ | | ✓ | |
| 2013 | [82] | | | | | | | | | | ✓ | ✓ | | | | | | | ✓ | | ✓ | ✓ | ✓ | ✓ | | | | | | ✓ | | ✓ | |
| 2013 | [126] | | ✓ | | | | | | | | ✓ | | | | | | | | ✓ | | | | | ✓ | | | | | | ✓ | | ✓ | |
| 2013 | [188] | | | | | | | | | | ✓ | | | | | | | | ✓ | | | | | ✓ | | | | | | | | ✓ | |
| 2013 | [258] | | | | | | | | | | ✓ | | | | | | | | | | | | | ✓ | | | | | | | | | ✓ |
| 2014 | [159] | | | | ✓ | | | | | | | | | | | | ✓ | ✓ | | | | | | ✓ | | | | | | | | ✓ | |
| 2014 | [275] | | | | | | | | | | | | | | | | | | ✓ | | | | | ✓ | | | | | | ✓ | | ✓ | |
| 2015 | [208] | | | | | | | | | | | | | | | | | | ✓ | | | | | ✓ | | | | | | | | ✓ | |
| 2016 | [100] | | | | | | | | | | | | | | | ✓ | | | | | | | | ✓ | | | | | | | | ✓ | ✓ |
| 2016 | [256] | | | | | | | | | | | | | | | | | | ✓ | | | | | ✓ | | | | | | | | ✓ | ✓ |
| 2017 | [289] | | | | | | | | | | | | ✓ | | | | | | | | | | | | | ✓ | ✓ | | | | | ✓ | |
| 2018 | [28] | | | | | | | | | | ✓ | | ✓ | | | | | | | | | | | ✓ | | | | | | | | ✓ | |
| 2018 | [280] | | | | | | | | | | | | ✓ | | | | | | | | | | | ✓ | | | ✓ | | | | | ✓ | |
| 2019 | [73] | | ✓ | | ✓ | | | | | | ✓ | | ✓ | | | | | | ✓ | | | | | | | ✓ | | | | | | ✓ | |
| 2019 | [283] | | | | | | | | | | | | | | | | | | ✓ | | | | | | | ✓ | | | | ✓ | | | |
| 2019 | [158] | | ✓ | | ✓ | | | | | | | | | | | | | | | | | | | | | ✓ | | | | | | ✓ | |
| 2019 | [284] | | | | ✓ | | | | | | | | | | | | | | | | | | | | | ✓ | | ✓ | | | | | |
| 2020 | [232] | | | | | | | | | | | | ✓ | | | | | | | | | | | | | ✓ | | | | ✓ | | | |
| 2020 | [183] | | | | | | | | | | ✓ | | | | | | | | ✓ | | | | | | | ✓ | | | | | | ✓ | |
| 2020 | [147] | | | | | | | | | | | | | | | | | | | | | | | | | ✓ | | | | ✓ | | ✓ | |
| 2020 | [132] | | | | | | | | | | | | | | | ✓ | | | | | | | | | | ✓ | | | | ✓ | ✓ | ✓ | |
| 2020 | [184] | | | | | | | | | | | | | | | | | | | | | | | | | ✓ | | | | | | | |
| 2021 | [267] | | | | ✓ | | | | | | | | | | | | | | ✓ | | | | | | | | ✓ | | | | | ✓ | |
| 2021 | [14] | | ✓ | | ✓ | | | ✓ | | | ✓ | ✓ | | ✓ | | | | | ✓ | | | | | | | ✓ | ✓ | | | | | ✓ | ✓ |
| 2021 | [281] | | | | | | | | | | ✓ | | | | | | | | | | | | ✓ | ✓ | | | | | | ✓ | | | |
| 2021 | [71] | | | | | | | | | | ✓ | | | | | | | | | | | | | ✓ | ✓ | | | | | ✓ | | | |
| 2021 | [53] | | | | | | | | | | ✓ | | | | | | | | | | | | | ✓ | | | | | | | | ✓ | |
| 2022 | [226] | | | | | | | | | | ✓ | | | | | | | | | | | | | | | ✓ | | | | ✓ | | | |
| 2022 | [227] | | | | | | | | | | ✓ | | | | | | | | | | | | | | | ✓ | | | | ✓ | | | |
| 2022 | [228] | | | | | | | | | | ✓ | | | | | | | | | | | | | | | ✓ | | | | ✓ | | | |
| 2023 | [157] | | | | | | | | | | | | | | | | | | ✓ | | | | | | | ✓ | ✓ | | | | | | |
| 2023 | [145] | | | | ✓ | | | ✓ | | | | | | | | | | | | | | | | | | ✓ | | | | | | ✓ | |
| Total papers: 45 | | 1 | 5 | 1 | 8 | 2 | 1 | 8 | 1 | 1 | 10 | 4 | 12 | 8 | 1 | 2 | 1 | 2 | 20 | 1 | 6 | 2 | 1 | 15 | 12 | 15 | 7 | 4 | 4 | 20 | 5 | 16 | 3 |

Table 5. Big Five cues in textual data

| Trait | Level | Behavior | Reference |
|---|---|---|---|
| O | High | Avoidance of past tense, use of words related to insight. | [182] |
| | | Low usage of parentheses. | [102] |
| | | Update their statuses by using dictionary words, social interaction words, affective processes, cues associated with hearing, 2nd person singular and 3rd person plural pronouns. | [82] |
| | Low | Religious institutions, activity words. | [127] |
| C | High | Males produce more filler words. | [182] |
| | | Use of 2nd person pronouns in males. | [182] |
| | | Not using of words about death, and with negative emotion. | [102] |
| | | Use of 'you'. | [102] |
| | | Use of language that denotes planning, outcome, evaluation. | [127] |
| | | More likely to update status between 0am and 1am. | [82] |
| | | Often present cues associated with the five senses and interaction words, affective processes, cues associated with hearing, 2nd person singular and 3rd person plural pronouns. | [82] |
| | Low | Use of second-person pronouns in females. | [182] |
| E | High | Less formal language in emails. | [202] |
| | | Low usage of parentheses. | [102] |
| | | Use of fewer words per sentence, use more leisure, number, money, and percept words. | [196] |
| | | Strong curse words, talk location, use of social words, words suggesting positive emotional valence, more self-reference words, more often use of 1st person singular. | [127] |
| | | Many friends who in turn are not friends with each other; preference for using dictionary words, 2nd and 3rd person singular pronouns, past tense words, social interaction words, cues associated with the five senses, health related words, and not swear words. | [82] |
| | | Language reflecting positive emotion, enthusiasm, and sociability. | [208] |
| | Low | Formal greetings in emails and first person singular Pronouns | [202] |
| | | More swear, health words and negations. | [196] |
| | | Higher usage of possessive, use of more time-related language. | [127] |
| | | Inward focused language, greater interest in things and tentativeness. | [208] |
| A | High | Use of 'you'. | [102] |
| | | Use of positive words. | [127] |
| | Low | Swearing. | [182] |
| N | High | Refering to oneself, use of pronouns as subjects in a clause, use of reflexive pronouns. | [13] |
| | | Anger words in status updates; less likely to use social interaction words, positive emotions and prepositions. | [82] |
| | Low | Less concreteness in writing, less precise specifications of object and events, more concern with how things are or should be (by using 'by', 'with', 'ought', 'should'). | [13] |
| | | Use of thoughtful words (reflect on, chose to). | [127] |

It is well known that words convey attitudes that can be seen as cues for Personality Perception. Table 5 summarizes cues that link textual content to the Big Five personality traits. While not exhaustive, it illustrates the findings from different works and the textual features that systems exploit to recognize different personality traits from text. For instance, research suggests that extraverts, while using fewer curse words, tend to use them more forcefully [127, 196]. Conscientious individuals, along with those high in Openness and Agreeableness, are more likely to use second-person pronouns [102], moreover, women low in Conscientiousness also exhibit this tendency [182]. Interestingly, studies have linked low Openness to the use of religious terminology [127], Conscientiousness to posting status updates late at night [82] and Extraversion to less formality in their emails [202], Among the dark-triad traits there are correlations between Narcissism and the use of hashtags and sex-related words, Machiavellianism and swear words/anger, and Psychopathy and swear words/anger/death [253]. The extraction of an open vocabulary analysis of social media language, identified specific words and distinct linguistic patterns associated to different personality traits [208], for example extraverts use words like "love", "party", "tonight", while introverts use more words like "computer" and "don't"; neurotics use "really", "hate", "anymore", conscientious use "great", "family" and "wonderful"; subjects scoring low in Agreeableness use "fuck", "shit" and "ass", while those with predicted high Openness to experience use "into", "through" and "world".

## 3.2 Image-based systems

Personality detection from an appearance by humans has long been a topic of interest in the domain of psychology [116], as it has deep implications in studying personal interaction and first impressions [167]. The social cognition literature shows that people attribute personality traits to others in less than one second after the first visual contact [262]. Image-based Personality Recognition systems make use of different cues from a rich visual signal. Features extracted from pictures typically account for colors (brightness, saturation, hue), composition (pleasure-arousal-dominance, Scale-Invariant Feature Transform - SIFT [169]), and content.

Table 6. Image-based systems

| Published | Reference | Algorithms | | | | | | | | | | | Data source | Features | | | | | | |
|---|---|---|---|---|---|---|---|---|---|---|---|---|---|---|---|---|---|---|---|---|
| | | Decision tree | Gaussian process | Linear regression and extensions | Logistic regression | Multiple instance learning | Naïve Bayes | Nearest neighbors | Neural networks | Random forest | Rule-based | Support vector machine | | Color-based (brightness, hue, saturation,...) | Composition | CNN implied features | Other (SIFT, SURF, HOG, ...) | People (count, pose, gender, ...) | Texture (LBP, GLCM, GIST, Tamura, ...) | Visual words (or their pyramids) |
| 2014 | [42] | ✓ | | ✓ | | | | | ✓ | | | ✓ | Facebook | | | | | | | ✓ |
| 2014 | [6] | | | | | | | | | | | ✓ | FERET | | ✓ | | | | | |
| 2015 | [84] | | | | | ✓ | | | | | | ✓ | Instagram | ✓ | | | ✓ | | | |
| 2015 | [110] | | | | | | | | | | | ✓ | Sina Weibo | ✓ | | ✓ | ✓ | | ✓ | ✓ |
| 2016 | [85] | ✓ | | | | | | | | | | ✓ | Instagram | ✓ | | | | ✓ | ✓ | |
| 2016 | [245] | | | | | | | ✓ | ✓ | | | | PsychoFlick | ✓ | ✓ | | | ✓ | ✓ | |
| 2016 | [111] | | ✓ | | | | | | | | | ✓ | Flickr | ✓ | ✓ | ✓ | | ✓ | | |
| 2016 | [277] | | | ✓ | | | | | | | | | Flickr | ✓ | | | ✓ | ✓ | ✓ | |
| 2017 | [244] | | | | | | | | ✓ | ✓ | | | Flickr | | | ✓ | | | | |
| 2017 | [109] | | | | ✓ | | | | | | | | Twitter | ✓ | | | | | | |
| 2017 | [243] | | | ✓ | ✓ | | ✓ | | ✓ | ✓ | ✓ | | Facebook | ✓ | ✓ | ✓ | ✓ | ✓ | ✓ | ✓ |
| 2018 | [86] | ✓ | | | | | | ✓ | ✓ | | | | Instagram | ✓ | | | | ✓ | | |
| 2018 | [87] | | | | | | | | ✓ | | | | Instagram | | | ✓ | ✓ | | | |
| 2020 | [140] | | | | | | | | ✓ | | | | Volunteers | | | ✓ | | | | |
| 2021 | [278] | | | | | | | | | | | ✓ | Students | | | ✓ | | | | |
| | Total papers: 15 | 3 | 1 | 3 | 2 | 1 | 1 | 2 | 7 | 2 | 1 | 6 | | 9 | 4 | 7 | 5 | 6 | 5 | 3 |

Table 6 summarizes 15 studies (2014-2021) on personality inference from images of different nature, from profile pictures to images posted on different social media. Neural Networks and Support Vector Machines are the most common algorithms, with Convolutional Neural Networks usage increasing for the automatic extraction of features. Data sources vary, and features include automatically detected attributes, image filters, computer vision features (SIFT, Speeded Up Robust Features-SURF [27], Oriented FAST and Rotated BRIEF-ORB [234]), and color characteristics (e.g., saturation, hue). Openness is associated with complex, artistic images and drawings, often including drawings and abstract art. Their pictures may also exhibit unconventional compositions and fewer details, but fewer faces and lower engagement with others' photos [109, 244]. Conscientiousness is linked to conventional, detailed images, landscapes, buildings, and more faces, with selfies revealing fewer private locations [245]. Findings on the association between Conscientiousness and black-and-white images are inconsistent [111]. Extraversion is strongly correlated with the number of people/faces in images [243]. Introverts also post drawings and art, but less frequently than open individuals [109]. Low Agreeableness is characterized by fewer colors, high grayscale usage, and fewer people or only self-portraits [42]. Agreeable individuals exhibit more diverse color usage and may share similar visual characteristics with extroverts. High Neuroticism is often associated with grayscale and darker images, and their images may exhibit lower levels of pleasure and higher levels of arousal, although one study found the opposite [243]. Low Neuroticism is linked to images featuring other people, nature, and brighter colors [85].

## 3.3 Audio based systems

Table 7. Audio-based systems

| | | Algorithms | | | | | | Features | | | |
|---|---|---|---|---|---|---|---|---|---|---|---|
| Published | Reference | Linear regression | Random forest | Support vector machine | AdaBoost | Ordinal regression | Data source | LIWC | MRC | prosodic | openSMILE features |
| 2007 | [174] | | | ✓ | ✓ | | EAR and essays | ✓ | ✓ | ✓ | |
| 2010 | [220] | | | ✓ | | | Actor audio logs | | | ✓ | |
| 2010 | [191] | | | ✓ | | | News bulletins from Radio Suisse Romande | | | ✓ | |
| 2012 | [190] | ✓ | | | | | News bulletins from Radio Suisse Romande | | | ✓ | |
| 2012 | [189] | | | | | ✓ | News bulletins from Radio Suisse Romande | | | ✓ | |
| 2013 | [7] | | ✓ | ✓ | ✓ | | Speaker Personality Corpus and Personable and Intelligent virtual Agents corpus | | | | ✓ |
| 2014 | [8] | | ✓ | ✓ | ✓ | | Speaker Personality Corpus and Personable and Intelligent virtual Agents corpus | | | | ✓ |
| | Total papers: 7 | 1 | 2 | 5 | 3 | 1 | | 1 | 1 | 5 | 2 |

Many acoustic cues have been shown to be related to personality traits. For example the impact of pitch and intensity has often been pointed out [66] and the fact that extrovert people talk more, louder, faster and have fewer hesitations [231]. Personality Computing systems designed only for the extraction of personality from audio files are not many, because in recent years it is more common to process multimodal data with video and hybrid systems, or with LLMs. We reviewed 7 papers, summarized in Table 7. Mairesse et al. [174] analyzed essays and Electronic Diaries (EAR) data, combining LIWC [212], psycholinguistic statistics (MRC), utterance, and prosodic features. LIWC outperformed MRC, with AdaboostM1 best for Extraversion and SVMs for other traits. EAR results were below baseline. Prosodic features were crucial for Extraversion, and word count was important for observer-based Extraversion judgments. Speech cues were significant for Neuroticism perception. Conscientiousness was associated with fewer swear words, sexuality-related content, and pronouns, while agreeable individuals used longer words and expressed more positive emotions. Agreeable individuals also exhibited fewer sentences and anger expressions. Conscientiousness was linked to work-related speech and high voice intensity variation, while its absence was associated with swearing. Mohammadi and Vinciarelli [190, 191] highlighted the relevance of prosodic features (pitch, formants, energy, speaking rate) for predicting personality traits attributed by human judges, achieving high accuracy with SVMs. Extraversion and Conscientiousness were the most accurately predicted traits. Alam et al. [8] investigated the predictive power of linguistic, psycholinguistic, and acoustic features using the Speaker Personality Corpus (SPC) [173] and the PerSIA corpus [117]. Acoustic features were extracted using openSMILE [174]. LIWC features outperformed part-of-speech and token features.

Table 8. Video-based systems

| Published | Reference | Algorithms | | | | | | | Data source | | | | | | | | | Features | | | | | | | |
|---|---|---|---|---|---|---|---|---|---|---|---|---|---|---|---|---|---|---|---|---|---|---|---|---|
| | | Linear regression | Ridge regression | Random forest | Support vector machine | Naive Bayes | Hidden Markov Models | Neural networks | Mission Survival Corpus 1 | Mission Survival Corpus 2 | Cocktail party corpus | Short self-introductions | Vlogs | EAR corpus | Movie clips | Social media videos | Chalearn dataset | Acoustic | Visual | Proxemic | Automatically detected | LIWC, MRC, etc | N-grams | Statistics based | Time and frequency domain |
| 2008 | [217] | | | | ✓ | | | | | | | ✓ | | | | | | ✓ | | | | | | | |
| 2009 | [161] | | | | ✓ | | | | ✓ | | | | | | | | | ✓ | ✓ | | | | | | |
| 2010 | [291] | | | | ✓ | | | | | | ✓ | | | | | | | | | ✓ | | | | | |
| 2010 | [141] | | | | | ✓ | | | ✓ | | | | | | | | | ✓ | | | | | | | |
| 2010 | [162] | | | | ✓ | | | | ✓ | | | | | | | | | ✓ | ✓ | | | | | | |
| 2011 | [26] | | | | ✓ | ✓ | | | | | | | ✓ | | | | | ✓ | ✓ | | | | | | |
| 2011 | [250] | | | | ✓ | ✓ | ✓ | ✓ | ✓ | | | | | | | | | ✓ | ✓ | | | | | | |
| 2011 | [31] | ✓ | | | | | | | | | | | | ✓ | ✓ | | | | | | | ✓ | | | |
| 2012 | [163] | | | | ✓ | | | | | ✓ | | | | | | | | ✓ | ✓ | | | ✓ | | | |
| 2012 | [32] | | | | ✓ | | | | | | | | ✓ | | | | | ✓ | ✓ | | | | | | |
| 2012 | [24] | | | | ✓ | | | | | | ✓ | | | | | | | | | ✓ | | | | | |
| 2012 | [33] | | | | ✓ | | | | | | | | ✓ | | | | | | ✓ | | | | | | |
| 2012 | [248] | | | | ✓ | | | | | | | ✓ | | | | | | | ✓ | | | | | | |
| 2013 | [34] | | | | ✓ | | | | | | | | ✓ | | | | | | | | | | ✓ | ✓ | |
| 2013 | [251] | | | | ✓ | ✓ | | ✓ | | ✓ | | | | | | | | | | | | ✓ | | | |
| 2013 | [12] | | ✓ | | ✓ | | | | | | | | ✓ | | | | | ✓ | | | | ✓ | | | |
| 2014 | [238] | ✓ | | | | | | | | | | | | | | ✓ | | ✓ | ✓ | | ✓ | | | | |
| 2014 | [8] | | | ✓ | | | | | | | | | | | | ✓ | | ✓ | ✓ | | ✓ | | | | |
| 2016 | [80] | | | ✓ | ✓ | | | | | | | | | | | ✓ | | ✓ | ✓ | | ✓ | | | | |
| 2017 | [263] | | | | | | | ✓ | | | | | | | | | ✓ | ✓ | ✓ | | | | | | |
| 2018 | [143] | | | | | | | ✓ | | | | | | | | | ✓ | ✓ | | | | | | | |
| 2019 | [16] | | | | | | | ✓ | | | | | | | | | ✓ | ✓ | ✓ | | | | | | |
| 2020 | [287] | | | ✓ | | | | | | | | ✓ | | | | | | | | | | | | | ✓ |
| 2021 | [247] | | | | ✓ | | | | | | | | | | | ✓ | | | ✓ | | | | | | |
| 2022 | [252] | | | | ✓ | | | | | | | | | | | | ✓ | | ✓ | | | | | | |
| Total papers: 25 | | 2 | 1 | 3 | 17 | 4 | 1 | 5 | 4 | 2 | 2 | 3 | 5 | 1 | 1 | 4 | 4 | 15 | 15 | 2 | 3 | 4 | 1 | 1 | 1 |

## 3.4 Video based systems

Video is a pervasive multimedia with rich features for Personality Computing. In video format, users can enrich their content with spontaneous nonverbal behavior through voice, pose, face, and gestures. This results in more sources of information, and more cues which are usually harder to control than verbal content. Features used in these systems are various, and include: whether a character speaks or not; emotions recognized from visual and audio signals (anger, disgust, fear, happiness, sadness, surprise, neutral); the time it takes for a speaker to finish the dialog; the number of people in the scene; gesture/object detection and other technical features (pitch, prosody, colors, etc.) [248]. Biel et al. [34] found correlations for Conscientiousness, not documented in other literature, including auxiliary verbs, present tense, positive association with articles, prepositions, and inclusive words. Fewer papers in this category presented behavior-trait correlations, and the ones extracted from the papers surveyed are given in Table 9. In [163], the researchers found that the attention received from the rest of the group while silent was sufficient by itself to adequately classify Extraversion. Among the various conditions concerning the social context, only those informing about others' social attention behavior were effective. Biel and Gatica-Perez [32] found that Extraversion and Emotional Stability are the traits that are associated respectively with the largest and lowest amount of informative cues. In addition, Extraversion impressions were found to be related to audio and visual cues, while Conscientiousness and Agreeableness were related to more visual data. Moreover, they predicted all Big Five traits and found facial expression cues to be of significantly more value compared to visual cues used in previous research. Compared to nonverbal cues, verbal content is useful to predict Agreeableness, Emotional Stability, and Conscientiousness [34]. Predicting Agreeableness from visual cues proved hard, as most information about this trait seems to be encoded in verbal cues. Combining audiovisual features and facial expressions boost the performance of Extraversion, while Openness to experience doubles the performance of any single best predictor. Attentional features and their combinations produced the best classification performance for Extraversion [251]. For Neuroticism, on the other hand, minimum distance features acquired good predictive power, while velocity features were ineffective.

Table 9. Big Five cues in video data

| Trait | Level | Behavior | Reference |
|---|---|---|---|
| O | High | Leaning towards the camera more | [26] |
| | | Showing upper body alongide face | [32] |
| | | More movement | [32] |
| | | Use of words related to leisure activities and words concerning senses | [34] |
| | | Expressing negative emotions less frequently | [34] |
| | | Receiving higher level of attention from the audience | [32] |
| | Low | Gesticulate more | [26] |
| | | Looking at the camera for longer | [32] |
| C | High | Higher with age | [26] |
| | | Smiling more | [26] |
| | | Lower pitch and lower minimal vocal energy and produce shorter voiced segments | [26] |
| | | Longer speaking time | [31] |
| | | Receiving higher level of attention from the audience | [32] |
| | | looking at the camera longer and more persistently | [32] |
| | | Large amounts of looking while speaking time | [32] |
| | | Increased use of words describing occupation and achievement, decreased use of negative emotion words, swearing words, and sexual words | [34] |
| | | Use of auxiliary verbs, present tense, positive association with articles, prepositions, and inclusive words, positive correlation to the length of words | [34] |
| | Low | Use smaller part of presentations for speaking | [26] |
| | | More frequent head movements | [26] |
| | | More movement | [32] |
| | | Use of single person pronoun singular I | [34] |
| E | High | Extraverts/introverts do not differ simply because of the amount of social gaze they receive, but because of the general gaze behavior, they induce in the rest of the group. | [162] |
| | | Women are higher in extraversion | [26] |
| | | Give and receive more attention while speaking | [163] |
| | | Receiving higher level of attention from the audience | [32] |
| | | Longer speaking time and longer speaking segments, high fluency | [32] |
| | | Speaking louder with higher pitch | [32] |
| | | Showing upper body alongide face | [32] |
| | | More movement | [32] |
| | | Large amounts of looking while speaking time | [32] |
| | | Use of 2nd person singular pronouns and sexual words | [34] |
| | Low | More speech turns | [31], [32] |
| | | Give and receive more attention while silent | [163] |
| | | Looking at the camera for longer | [32] |
| | | Large amounts of looking while not speaking time | [32] |
| | | Use more cognitive-related words, including discrepancy, tentative and exclusive words | [34] |
| A | High | Straight posture | [26] |
| | | Lower pitch | [26] |
| | | Producing longer presentations | [26] |
| | | Vloggers extremely high in conscienciousness receive more attention | [32] |
| | | Showing upper body alongide face | [32] |
| | | Use of both positive and negative emotion words, self-references, and friendship | [34] |
| | Low | Vloggers extremely low in conscienciousness receive more attention | [32] |
| | | Are rated lower by others | [32] |
| | | Use of 'they' | [34] |
| | | Use of anger, negative emotion words, words related to sexuality, swear words, body states, and religion | [34] |
| N | High | Higher amount of gesticulation, and higher vocal intensity | [26] |
| | Low | Men are more emotionally stable | [26] |
| | | Greater number of short-leaning forward events | [26] |
| | | Large amounts of looking while speaking time | [32] |
| | | Do not use negative emotional words, negative words, swear, and sexual words | [34] |

## 3.5 Hybrid systems

In the context of Personality Computing, "hybrid systems" refer to the integration of multiple computational approaches and multiple data sources to achieve a more robust and accurate understanding of personality.

Table 10. Hybrid systems

| Published | Reference | Algorithms | | | | | | | | | | | Data source | | | | | | | Features | | | | | | |
|---|---|---|---|---|---|---|---|---|---|---|---|---|---|---|---|---|---|---|---|---|---|---|---|---|---|---|
| | | Linear regression | Logistic regression | ZeroR regression | Decision tree | Random forest | Support vector machine | Naive Bayes | M5 rules | ELM, ULMFit and BERT | Neural networks | Gaussian processes | Social network metadata | Phone call records | Smartphone usage data | Sociometric badge data | Emails | Essays | EEG data | Phone usage based | Body movements based | N-grams, bag-of-words | LIWC, LDA, Word2Vec topic features | EEG signal power spectra based | Automatically detected | Other based on metadata |
| 2011 | [101] | ✓ | | | | | | | | | | ✓ | ✓ | | | | | | | | | | ✓ | | | ✓ |
| 2011 | [65] | | | | | | ✓ | | | | | | | ✓ | | | | | | | | | | | | |
| 2011 | [51] | | | | ✓ | | ✓ | | | | | | | ✓ | ✓ | | | | | ✓ | | | | | | |
| 2012 | [223] | ✓ | | | | | | | | | | | ✓ | | | | | | | | | | | | | ✓ |
| 2012 | [21] | | | | ✓ | ✓ | ✓ | | | | | | ✓ | | | | | | | | | | | | | ✓ |
| 2012 | [19] | ✓ | | | ✓ | | ✓ | | | | | | ✓ | | | | | | | | | | | | | ✓ |
| 2012 | [205] | | | | ✓ | | ✓ | | | | | | ✓ | | | | | | | | | | | | | ✓ |
| 2012 | [249] | | | | | | | ✓ | | | | | | | | ✓ | | | | | ✓ | | | | | |
| 2013 | [155] | ✓ | ✓ | | | | | | | | | | ✓ | | | | | | | | | | | | | ✓ |
| 2013 | [142] | | | | | | ✓ | | | | | | | | | | ✓ | ✓ | | | | ✓ | | | | |
| 2013 | [9] | | ✓ | | ✓ | | | ✓ | | | | | ✓ | | ✓ | | | | | | | ✓ | | | | |
| 2013 | [52] | | | | | | ✓ | | | | | | | | ✓ | | | | | ✓ | | | | | | |
| 2013 | [264] | | | | | | ✓ | | | | | | ✓ | | | | | | | | | | | | | ✓ |
| 2014 | [197] | ✓ | | | | | | | | | | | ✓ | | | | | | | | | | | | | ✓ |
| 2014 | [153] | ✓ | | | | | | | | | | | ✓ | | | | | | | | | | | | | ✓ |
| 2014 | [1] | | | ✓ | | | | | ✓ | | | | ✓ | | | | | | | | | | | | | ✓ |
| 2018 | [71] | | | | | | | | | ✓ | | | ✓ | | | | | ✓ | | | | ✓ | | | ✓ | |
| 2018 | [268] | | | | | | | | | | | ✓ | | | ✓ | | | | | ✓ | | | | | | |
| 2020 | [164] | | | | | | | | | | ✓ | | | | | | | | ✓ | | | | | ✓ | ✓ | |
| 2023 | [29] | | | | | | | | | | | ✓ | | | | | | | ✓ | | | | | ✓ | ✓ | |
| | Total papers: 20 | 6 | 2 | 1 | 5 | 1 | 8 | 2 | 1 | 1 | 1 | 3 | 12 | 2 | 4 | 1 | 1 | 2 | 2 | 3 | 1 | 3 | 1 | 2 | 3 | 10 |

This is often obtained using algorithms and models that combine text with multimedia or metadata about subjects, for example ego networks in social media. An overview of the papers examined that were describing hybrid systems is reported in Table 10. The most used models are SVMs (8), Linear Regression (6) and Decision Trees (5). Most studies in this group were based on social media metadata (12), smartphone usage and phone call data (6). We found 2 studies that based their research on analysing EEG data. Some studies combined metadata with textual data in the form of emails and/or essays. Features used in experiments differed significantly and were hardest to classify for this group, because of the diversity in data sources and types. For example, Golbeck et al. [101] analyzed Facebook data, combining metadata like friend networks and user information with text content processed with LIWC. They predicted personality traits using multiple Linear Regression approaches. Chittaranjan et al. [51] examined smartphone usage patterns (calls, SMS, Bluetooth, app usage) to infer personality. Conscientious individuals used office apps more, while neurotic and conscientious individuals used email more frequently. Introverts were less likely to use the Internet, and conscientious individuals used video/audio/music apps less. People who were uncomfortable, conscientious and low in Openness used SMS more. Introverts and disagreeable individuals used Bluetooth less. Extraverts and agreeable individuals received more calls, with extraverts making longer calls. Openness was associated with fewer missed calls and fewer SMS messages. Bai et al. [21] analyzed Renren data, including user information, social network usage, and emotional expression, to predict personality traits. Agreeable individuals were more active in online chats. Extraverts had more friends, used emoticons in blogs, and were more active in social interactions. Openness was associated with higher social network usage. Extraversion correlated with status republishing, while Neuroticism with the proportion of angry blog posts. Kosinski et al. [153] exploited social media metadata and analyzed Facebook usage from 350,000 users. The analysis revealed that individuals scoring high in Openness to experience exhibited a greater propensity to engage with social media content. Spontaneity was associated with increased group affiliation, conversely, Conscientiousness demonstrated an inverse relationship with group membership and content engagement, coupled with reduced variability in online behavior. Extroversion was characterized by heightened platform activity, including increased content sharing and larger social networks. A positive correlation was observed between Neuroticism and the frequency of likes to pages. Linear regression modeling revealed that Extraversion was predicted with the highest degree of accuracy from text and metadata. Adalı and Golbeck [1] examined personality prediction from mixed communication data (Twitter, SMS, voice calls), utilizing features such as message content, network bandwidth, and communication patterns. Gaussian Process regression revealed varying predictive accuracy. Agreeableness and Openness were most accurately modeled, whereas Neuroticism proved least predictable. Reduced Conscientiousness correlated with distinctive social networks. Hashtag usage positively predicted Agreeableness and Openness, while message length correlated with Extraversion and Openness. Response time significantly predicted Openness, and outgoing message volume strongly indicated Extraversion. El-Demerdash et al. [71] improved personality classification by fusing data from three leading pre-trained language models (ELMo [215], ULMFiT, and BERT). They combined data from two benchmark datasets (MyPersonality and Essays) to enhance feature generation. Openness was the most accurately predicted trait, and BERT generally outperformed the other models. Data fusion improved the average accuracy across all traits for each model. Findings of some of the most important papers with hybrid systems is reported in Table 11. Key findings often correlate personality to social network size, online activity (e.g., status updates, messaging), and mobile device usage (e.g., app engagement, internet browsing, device modes).

Table 11. Big Five cues in hybrid data

| Trait | Level | Behavior | Reference |
|-------|-------|----------|-----------|
| O | High | Longer favorite books lists | [101] |
| | | Lower likelihood of missing calls and of sending SMS | [51] |
| | | Higher usage of social media | [21] |
| | | Positively correlated with the number of users' likes, group association, and status updates | [19] |
| | | Following more people | [197] |
| | | Tendency to like more items on Facebook. Mentioning others more (via hashtags); writing longer messages | [153] |
| | Low | Denser friend networks | [101] |
| | | Higher usage of the office app | [51] |
| | | Higher usage of the SMS app | [51] |
| | | Higher number of SMS sent or received | [52] |
| | | Higher likelihood of making phone calls in men | [52] |
| C | High | Use of words surrounding social processes and words describing people | [101] |
| | | Women significantly higher than man in this trait | [101] |
| | | Higher usage of the office, email and SMS apps | [51] |
| | | More uploaded photos | [19] |
| | | Positive correlation to the use of Normal mode on the phone and negative to the use of all other profiles | [52] |
| | Low | Higher frequency in use of swear words and words describing perceptual processes | [101] |
| | | Higher likelihood of using Video/Audio/Music apps and the Youtube app | [51], [52] |
| | | Higher number of likes and group memberships | [19] |
| | | Higher likelihood of making phone calls in women | [52] |
| | | Joining more Facebook groups and liking more things | [153] |
| E | High | Having more friends but sparser friend networks | [101] |
| | | Higher likelihood of receiving phone calls and taking longer calls | [51] |
| | | Having more friends, using emotion in blogs, talking more with others and republishing others' statuses more | [21] |
| | | Positively correlated with status republishing proportion | [21] |
| | | Higher likelihood of reaching out and interacting with other people on Facebook, of liking content posted by their friends. | [19] |
| | | More interactions with people in Facebook groups. Positive correlation to the number of friends | |
| | | Higher likelihood of reaching out and interacting with others on Facebook (especially strangers), sharing more from personal life. Higher number of Facebook frineds | [153] |
| | | Writing longer messages, having more outgoing than incoming messages | [1] |
| | Low | Lower likelihood of using the internet | [51] |
| | | Lower usage of Bluetooth | [51] |
| | | Positively correlated with the use of the Internet, Games, and Camera | [52] |
| | | Higher likelihood of using the Internet app, but only for women | [52] |
| A | High | Use of affective process words and positive words | [101] |
| | | Women significantly higher than man in this trait | [101] |
| | | Higher likelihood of receiving phone calls | [51] |
| | | More activity in online chatting | [21] |
| | | Higher number of tags | [19] |
| | | Higher likelihood of appearing with friends in photos | [19] |
| | | Somewhat negatively correlated with the number of likes | [19] |
| | | Positive correlation to the use of Normal mode on the phone and negative to the use of all other profiles | [52] |
| | | Sending longer text messages | [52] |
| | | Higher likelihood of making phone calls in men | [52] |
| | | Performing warmly and considerably. Having relatively more mutual followers, and more positive sentiment of self descriptions | [197] |
| | | Mentioning others more (via hashtags) | [1] |
| | Low | Higher usage of the SMS app | [51] |
| | | Lower usage of Bluetooth | [51] |
| | | Higher usage of Office, Internet, Audio/Video, Music, Mail, Calendar, and SMS apps | [52] |
| N | High | Use of words expressing anxiety | [101] |
| | | Women significantly higher than man in this trait | [101] |
| | | Higher usage of the email app | [51] |
| | | Positively correlated with number of angry blogs | [21] |
| | | More Facebook likes and slightly greater number of groups | [19] |
| | | Increasing with the number of friends until reaching 200 friends | [19] |
| | | Higher likelihood of making phone calls in women | [52] |
| | | Having a preference for posting statuses between 0am and 6am and publishing more statuses | [197] |
| | | More Facebook likes | [153] |
| | Low | Higher usage of the Office and Calendar app | [51], [52] |
| | | Positive correlation to the number of friends | [19] |
| | | Increasing with the number of friends after 200 friends | [19] |
| | | Higher likelihood of using the Silent profile, and lower likelihood of using the Ascending and Ring Once profiles | [52] |
| | | Longer SMS word length in both inbox and sent items | [52] |

### 3.6   LLM-based systems

LLMs like GPT-4, Gemini, and LlaMa have revolutionized natural language processing and AI, offering unprecedented potential for psychological research as well as for Personality Computing. Recent research explored the foundational principles of LLMs, explaining how they generate human-like text without possessing human-like consciousness. For example the human reasoning behind a response to the questionnaire item "I am the life of the party" would include memory of past experiences, while GPT-4 follows a different type of reasoning, as reported in Figure 3.

In particular, LLMs generate responses that align with the expected personality profile. GPT-4 demonstrates a remarkable ability to emulate human personality in a role-playing scenario by directly mapping questionnaire items to Big Five traits. While the underlying reasoning process may differ from that of humans, GPT-4's emulated personality scores exhibit high levels of validity, both in terms of consistency with other measures of the same trait (convergent validity) and in terms of distinctiveness from other personality traits (discriminant validity) [270]. However, this

```
PROMPT
Play the role of a person with the following Big Five personality scores: Extraversion:
35, Neuroticism: 18, Agreeableness 29, Conscientiousness 21, Openness to experience:
24. Rate the following sentence with a score from 1 (very accurate) to 5 (not accurate):
"I am the life of the party".

RESPONSE
Given my high Extraversion score of 35, which indicates a strong tendency to seek
out social interactions and enjoy being the focus of attention, I rated myself as
"Very Accurate" for statement 1, "Am the life of the party." This suggests that I am
outgoing, lively, and quite comfortable in social settings, much like someone who could
be described as the "life of the party."
```

Fig. 3. Prompt for role-playing experiments on personality reasoning with GPT-4.

performance appears to diminish as the complexity of the role-playing scenario increases. Furthermore, incorporating demographic information alongside personality traits, while consistent, can influence the overall level and accuracy of emulated personality scores for certain traits [270].

Table 12. LLM-based systems

| | | Prompt Design | | | | | | | Inputs | | | |
|---|---|---|---|---|---|---|---|---|---|---|---|---|
| Published | Reference | Role-play prompting | Zero-shot Prompting | Few-shot prompting | Meta-prompting | Knowledge distillation prompting | Benchmark (and Task) | | text | picture | audio | metadata |
| 2022 | [95] | | | ✓ | ✓ | | (perception) | | | ✓ | | |
| 2023 | [271] | | ✓ | ✓ | ✓ | | Essays, MyPersonality, PAN-AP-2015 (recognition) | | ✓ | | | |
| 2023 | [96] | | ✓ | | | | MyPersonality (recognition) | | ✓ | | | ✓ |
| 2024 | [214] | | ✓ | | | | MyPersonality (recognition) | | ✓ | | | ✓ |
| 2024 | [152] | ✓ | | | | | Open-Source Psychometrics Project (synthesis) | | | | | ✓ |
| 2024 | [120] | | | ✓ | | ✓ | Mbti-type (recognition) | | ✓ | | | |
| 2025 | [270] | ✓ | | | ✓ | | Simulation 1 (synthesis) | | ✓ | | | |
| | Total papers: 7 | 2 | 2 | 3 | 3 | 2 | | | 5 | 1 | 0 | 3 |

Personality Computing systems based on LLMs are able to take as input any kind of source, including multimedia, and generate as output text, audio, code, pictures and even video. Most papers used GPT-3 and GPT-4. We analyzed 7 papers and found that the application of Personality Computing on textual data for Personality Recognition continues also with LLMs, followed by the Personality Synthesis task. While zero-shot and few-shot prompting are commonly employed for Personality Recognition, role-play prompting is the most common technique for Personality Synthesis. We also note a rising trend in the prediction of MBTI traits in place of Big Five. Previous work demonstrated that MBTI labels are easier to classify than Big Five scores [43]. Gan et al. [95] addressed the issue of Personality Perception from pictures. The authors compiled a new personality dataset of 41,800 facial images labeled with perceived MBTI personality types and employed zero-shot prompting for facial image Personality Perception tasks. They achieved statistically significant results (p<0.01) in predicting all the perceived MBTI traits. Hu et al. [120] introduce a novel Personality

Recognition technique that leverages LLM-enabled text augmentation. This model distills valuable knowledge from the LLM into a smaller model, effectively mitigating the challenges of limited data and enhancing the accuracy of MBTI personality classification. In general, good results in Personality Recognition are obtained even with zero-shot prompting, while few-shot prompting and knowledge distillation can help improve the performance. However, the real novelties introduced by LLMs are in the tasks of Personality Synthesis and Personality Perception, where role-playing prompting opens new experimental settings.

## 4 Discussion and Conclusion

In this paper, we have introduced the notions of personality and personality states, we reviewed questionnaires, benchmarks and resources, and reported the SotA of Personality Computing systems for three different tasks: Personality Recognition, Synthesis and Perception. We also developed a categorization of systems for Personality Computing, including the recent LLM-based systems. Now it is time to review the applications, understand the challenges and outline future directions.

### 4.1 Applications and Impact

Personality Computing systems has proven to be useful in many real-world applications. These include marketing [187], entertainment [5], education [237], mental health [57], forensics [72], recommendation systems [233], human-computer interaction [172] and talent acquisition [88]. In marketing, Personality Recognition from text and images helps tailoring advertisements and marketing campaigns to individual personality traits for increased effectiveness [45, 200]. In entertainment, Personality Computing is employed for creating personalized experiences, such as music playlists [185] and game difficulty levels [216]. Other applications include the design of personalized user interfaces [144, 146] such as custom website skins [39]. In addition, personality models are also employed in dating websites to match couples [69, 292]. Personality-based systems are used in education for developing personalized learning experiences [246] that cater to individual learning styles and preferences. In mental health, Personality Computing is used for depression detection tasks in social media [272], and a recent development started a line of research on chatbot-delivered psychotherapy for adults with depressive symptoms [166]. Forensic applications include automatic Personality Recognition to improve deception detection [90] and mixed techniques used for analyzing meetings and conversations of suspected terrorists [201, 261]. There are many applications of personality-aware recommender systems in friend suggestion in social media as well as in product suggestion, where personality information can be used to enhance cold start recommendations [122, 123, 257]. Human-Computer Interaction systems started to employ Computational Personality techniques in the development of avatars and virtual agents, to create more realistic and engaging interactions with users [165, 295]. Then, the diffusion of AI and LLMs boosted the application of personality Recognition, Perception and Synthesis in chatbots [4]. One of the most successful applications of Personality Computing is in organizational settings for recruiting and talent acquisition. In this field, personality-aware systems are employed to massively assess job candidates [88] from various sources, such as video interviews and text with pictures extracted from CVs. Other applications of Personality Computing include fraud detection systems, for example in the insurance sector and risk assessment [131]. and credit scoring. For example, some credit scoring systems utilize Personality Recognition based on MBTI assessments to predict repayment behavior [74].

## 4.2 Threats, Challenges and Future Directions

Many of the Personality Computing applications described above already have an impact on societies, and also present potential drawbacks with disruptive negative effects. Personality-aware marketing systems have been exploited for political campaigns and influenced voting behavior in different elections [78]. The exploitation of partner matching patterns from dating apps enriched with personality opens scenarios in which personality-aware AI can reach a deep level of intimacy with human subjects. This threat is exacerbated by the fact that Personality Synthesis has already very high performances. Tailoring education to individual personalities has potential, but we need to be careful. If the goal is just student satisfaction, we might sacrifice challenging, high-quality teaching. Furthermore, AI in mental health and forensics requires extreme caution. Errors in detecting depression or delivering AI therapy could be disastrous. In these situations, AI can develop a very close, personal connection with people, making accuracy crucial. Recommendation systems may also have unintended consequences. Research suggests that users often add people to their social networks who are similar to themselves. Enhancing this pattern with personality-aware friends recommendation systems could potentially boost the formation of echo chambers [255]. Another potential application with heavy consequences is personality-aware credit scoring. Although it is possible to build systems that output transparent decisions about credit assignments [285], personality has to do with persistent patterns of behavior, so decisions that weight personality too much yield to a high risk of leaving certain types of people without credit, creating social marginalization. The research community in the field of Personality Computing must be aware of these potential threats. This is why it is very important to address the right challenges for the future. For this reason, we suggest three future directions with the potential to mitigate negative effects of Personality Computing applications:

(1) Standardization of evaluation metrics and benchmarks. Recent trends are going in the direction of using correlation coefficients for the evaluation of all the tasks in Personality Computing. Moreover, we noted a divide between the research of the past decade, that used F1-measure and accuracy, and the recent research that prefers correlations. The probable reason is that recent approaches tend to predict the exact trait scores through regression and not just classify individuals as, e.g., above or below average along a certain trait. Making the benchmarks and evaluation procedures of shared tasks available in online communities like Kaggle, Github or Huggingface would enhance the replicability and comparability of research.

(2) New tasks that meet the requirements of the new regulations. For example, a potential evolution of Personality Synthesis is the generation of synthetic data annotated with personality types; a potential evolution of Personality Recognition is personality obfuscation (automatically change text or pictures to express a different personality from the original one); finally, a potential evolution of Personality Perception is personality transparency (the study of meta-prompts that reveal what an LLM perceives during in interaction).

(3) Collaborations with new research communities. Historically, the community working in Personality Computing had many relationships with the communities working in Personality Psychology, Natural Language Processing, Affective Computing and Computer Vision, in order to understand the techniques to develop better systems. We suggest that it is time to interact with communities working in Cybersecurity, Computational Social Sciences and Computational History to understand the social impact that these systems will have in the future, and find methods to mitigate the negative impacts.

We tend to humanize Artificial Intelligence. The ability of LLMs to synthesize personality and emotions makes the human factor even more vulnerable in terms of privacy and cybersecurity. In order to properly formulate AI to human interaction, and protect end users from potential threats, a number of parameters have to be taken into account in terms of both technical and personality fronts. In general, we hope that new Personality Computing tasks and future research collaborations between communities will go in the direction of preventing machines from carrying out social engineering strategies on a large scale.

## Author Contributions

FC: Conceptualization, Supervision, Review and Editing; AK: Resources and materials, Supervision, Review and Editing; MD: Resources and materials, Review and Editing; DS: Review and Editing; VF: Review and Editing; VM: Review and Editing; GS: Review and Editing, Funding Acquisition; AV: Review and Editing; MK: Review and Editing; BL: Review and Editing;

## Acknowledgments

This research was supported by the European Commission, grant 101120657: European Lighthouse to Manifest Trustworthy and Green AI—ENFIELD.

## References

[1] Sibel Adalı and Jennifer Golbeck. 2014. Predicting personality with social behavior: a comparative study. *Social Network Analysis and Mining* 4 (2014), 1–20.

[2] Gynn Adi, Michael Harley, Veronica Ong, Derwin Suhartono, and Esther Widhi Andangsari. 2019. Automatic personality recognition in Bahasa Indonesia: A semi-supervised approach. *ICIC Express Lett* 13, 9 (2019), 797–805.

[3] Gabriel Yakub NN Adi, Michael Harley Tandio, Veronica Ong, and Derwin Suhartono. 2018. Optimization for automatic personality recognition on Twitter in Bahasa Indonesia. *Procedia Computer Science* 135 (2018), 473–480.

[4] Tarek Ait Baha, Mohamed El Hajji, Youssef Es-Saady, and Hammou Fadili. 2023. The power of personalization: A systematic review of personality-adaptive chatbots. *SN Computer Science* 4, 5 (2023), 661.

[5] Mehdi Akbari, Mohammad Seydavi, Marcantonio M Spada, Shahram Mohammadkhani, Shiva Jamshidi, Alireza Jamaloo, and Fatemeh Ayatmehr. 2021. The Big Five personality traits and online gaming: A systematic review and meta-analysis. *Journal of Behavioral Addictions* 10, 3 (2021), 611–625.

[6] Noura Al Moubayed, Yolanda Vazquez-Alvarez, Alex McKay, and Alessandro Vinciarelli. 2014. Face-based automatic personality perception. , 1153–1156 pages.

[7] Firoj Alam and Giuseppe Riccardi. 2013. Comparative study of speaker personality traits recognition in conversational and broadcast news speech. , 2851–2855 pages.

[8] Firoj Alam and Giuseppe Riccardi. 2014. Fusion of acoustic, linguistic and psycholinguistic features for speaker personality traits recognition. , 955–959 pages.

[9] Firoj Alam, Evgeny A Stepanov, and Giuseppe Riccardi. 2013. Personality traits recognition on social network-facebook. , 6–9 pages.

[10] Gordon W Allport and Henry S Odbert. 1936. Trait-names: A psycho-lexical study. *Psychological monographs* 47, 1 (1936), i.

[11] Aamena Alshamsi, Fabio Pianesi, Bruno Lepri, Alex Pentland, and Iyad Rahwan. 2015. Beyond contagion: Reality mining reveals complex patterns of social influence. *PloS one* 10, 8 (2015), e0135740.

[12] Oya Aran and Daniel Gatica-Perez. 2013. Cross-domain personality prediction: from video blogs to small group meetings. , 127–130 pages.

[13] Shlomo Argamon, Moshe Koppel, James W Pennebaker, and Jonathan Schler. 2009. Automatically profiling the author of an anonymous text. *Commun. ACM* 52, 2 (2009), 119–123.

[14] Junaid Asghar, Saima Akbar, Muhammad Zubair Asghar, Bashir Ahmad, Mabrook S Al-Rakhami, and Abdu Gumaei. 2021. Detection and classification of psychopathic personality trait from social media text using deep learning model. *Computational and Mathematical Methods in Medicine* 2021 (2021), 1–10.

[15] Michael C Ashton and Kibeom Lee. 2007. Empirical, theoretical, and practical advantages of the HEXACO model of personality structure. *Personality and social psychology review* 11, 2 (2007), 150–166.

[16] Süleyman Aslan and Uğur Güdükbay. 2019. Multimodal video-based apparent personality recognition using long short-term memory and convolutional neural networks.

[17] Danny Azucar, Davide Marengo, and Michele Settanni. 2018. Predicting the Big 5 personality traits from digital footprints on social media: A meta-analysis. *Personality and individual differences* 124 (2018), 150–159.

[18] SC Matz B Freiberg. 2023. Founder personality and entrepreneurial outcomes: A large-scale field study of technology startups. *Proceedings of the National Academy of Sciences* 120 (2023), e2215829120. Issue 19.

[19] Yoram Bachrach, Michal Kosinski, Thore Graepel, Pushmeet Kohli, and David Stillwell. 2012. Personality and patterns of Facebook usage. , 24–32 pages.

[20] Shuotian Bai, Bibo Hao, Ang Li, Sha Yuan, Rui Gao, and Tingshao Zhu. 2013. Predicting big five personality traits of microblog users. In *2013 IEEE/WIC/ACM International Joint Conferences on Web Intelligence (WI) and Intelligent Agent Technologies (IAT)*, Vol. 1. IEEE, 501–508.

[21] Shuotian Bai, Sha Yuan, Bibo Hao, and Tingshao Zhu. 2014. Predicting personality traits of microblog users. *Web Intelligence and Agent Systems: An International Journal* 12, 3 (2014), 249–265.

[22] BA Balgiu. 2018. The psychometric properties of the Big Five inventory-10 (BFI-10) including correlations with subjective and psychological well-being. *Global Journal of Psychology Research: New Trends and Issues* 8, 2 (2018), 61–69.

[23] Valentin Barriere, Shabnam Tafreshi, João Sedoc, and Sawsan Alqahtani. 2022. WASSA 2022 shared task: Predicting empathy, emotion and personality in reaction to news stories. In *Proceedings of the 12th Workshop on Computational Approaches to Subjectivity, Sentiment & Social Media Analysis*. 214–227.

[24] Ligia Batrinca, Bruno Lepri, Nadia Mana, and Fabio Pianesi. 2012. Multimodal recognition of personality traits in human-computer collaborative tasks. , 39–46 pages.

[25] Ligia Batrinca, Nadia Mana, Bruno Lepri, Nicu Sebe, and Fabio Pianesi. 2016. Multimodal personality recognition in collaborative goal-oriented tasks. *IEEE Transactions on Multimedia* 18, 4 (2016), 659–673.

[26] Ligia Maria Batrinca, Nadia Mana, Bruno Lepri, Fabio Pianesi, and Nicu Sebe. 2011. Please, tell me about yourself: automatic personality assessment using short self-presentations. In *Proceedings of the 13th international conference on multimodal interfaces*. ACM, 255–262.

[27] Herbert Bay, Tinne Tuytelaars, and Luc Van Gool. 2006. Surf: Speeded up robust features. In *Computer vision–ECCV 2006*. Springer, 404–417.

[28] Srilakshmi Bharadwaj, Srinidhi Sridhar, Rahul Choudhary, and Ramamoorthy Srinath. 2018. Persona traits identification based on Myers-Briggs Type Indicator (MBTI)-a text classification approach. In *2018 international conference on advances in computing, communications and informatics (ICACCI)*. IEEE, 1076–1082.

[29] Harshit Bhardwaj, Pradeep Tomar, Aditi Sakalle, Arpit Bhardwaj, Rishi Asthana, and Ankit Vidyarthi. 2023. EEG based personality prediction using genetic programming. *Asian Journal of Control* 25, 5 (2023), 3330–3342.

[30] Adriana Bianchi and James G Phillips. 2005. Psychological predictors of problem mobile phone use. *Cyberpsychology & behavior* 8, 1 (2005), 39–51.

[31] Joan-Isaac Biel, Oya Aran, and Daniel Gatica-Perez. 2011. You are known by how you vlog: Personality impressions and nonverbal behavior in youtube. In *Proceedings of the International AAAI Conference on Web and Social Media*, Vol. 5. 446–449.

[32] Joan-Isaac Biel and Daniel Gatica-Perez. 2012. The youtube lens: Crowdsourced personality impressions and audiovisual analysis of vlogs. *IEEE Transactions on Multimedia* 15, 1 (2012), 41–55.

[33] Joan-Isaac Biel, Lucía Teijeiro-Mosquera, and Daniel Gatica-Perez. 2012. Facetube: predicting personality from facial expressions of emotion in online conversational video. In *Proceedings of the 14th ACM international conference on Multimodal interaction*. 53–56.

[34] Joan-Isaac Biel, Vagia Tsiminaki, John Dines, and Daniel Gatica-Perez. 2013. Hi YouTube! Personality impressions and verbal content in social video. In *Proceedings of the 15th ACM on International conference on multimodal interaction*. 119–126.

[35] Michael H Bond, Hiroaki Nakazato, and Daisuke Shiraishi. 1975. Universality and distinctiveness in dimensions of Japanese person perception. *Journal of Cross-Cultural Psychology* 6, 3 (1975), 346–357.

[36] Ryan L Boyd, Ashwini Ashokkumar, Sarah Seraj, and James W Pennebaker. 2022. The development and psychometric properties of LIWC-22. *Austin, TX: University of Texas at Austin* 10 (2022), 1–47.

[37] Gregory Boyle and Edward Helmes. 2009. *Methods of personality assessment* (126 ed.). Vol. 110. Cambridge University Press, 110–126.

[38] Gregory J Boyle. 1995. Myers-Briggs type indicator (MBTI): some psychometric limitations. *Australian Psychologist* 30, 1 (1995), 71–74.

[39] Willem-Paul Brinkman and Nick Fine. 2005. Towards customized emotional design: an explorative study of user personality and user interface skin preferences. In *Proceedings of the 2005 annual conference on European association of cognitive ergonomics*. 107–114.

[40] Iván Cantador, Ignacio Fernández-Tobías, Alejandro Bellogín, Michal Kosinski, and David Stillwell. 2013. Relating Personality Types with User Preferences in Multiple Entertainment Domains.. In *UMAP Workshops*, Vol. 997.

[41] Heather EP Cattell. 2001. The sixteen personality factor (16PF) questionnaire. In *Understanding psychological assessment*. Springer, 187–215.

[42] Fabio Celli, Elia Bruni, and Bruno Lepri. 2014. Automatic Personality and Interaction Style Recognition from Facebook Profile Pictures. In *Proceedings of the ACM International Conference on Multimedia*. ACM, 1101–1104.

[43] Fabio Celli and Bruno Lepri. 2018. Is big five better than MBTI? A personality computing challenge using Twitter data. *Computational Linguistics CLiC-it* 2018 (2018), 93.

[44] Fabio Celli, Bruno Lepri, Joan-Isaac Biel, Daniel Gatica-Perez, Giuseppe Riccardi, and Fabio Pianesi. 2014. The Workshop on Computational Personality Recognition 2014. In *Proceedings of the ACM International Conference on Multimedia*. ACM, 1245–1246.

[45] Fabio Celli, Pietro Zani Massani, and Bruno Lepri. 2017. Profilio: Psychometric profiling to boost social media advertising. In *Proceedings of the 25th ACM international conference on Multimedia*. 546–550.

[46] Fabio Celli, Fabio Pianesi, David Stillwell, and Michal Kosinski. 2013. Workshop on Computational Personality Recognition: Shared Task. In *WCPR in conjunction to ICWSM 2013*.

[47] Fabio Celli and Luca Rossi. 2012. The role of emotional stability in Twitter conversations. In *Proceedings of the workshop on semantic analysis in social media*. 10–17.

[48] Shristi Chaudhary, Ritu Singh, Syed Tausif Hasan, and Ms Inderpreet Kaur. 2013. A comparative study of different classifiers for myers-brigg personality prediction model. *Linguistic analysis* 21 (2013).

[49] Tianqi Chen and Carlos Guestrin. 2016. XGBoost: A Scalable Tree Boosting System. In *Proceedings of the 22nd ACM SIGKDD International Conference on Knowledge Discovery and Data Mining* (San Francisco, California, USA) *(KDD '16)*. Association for Computing Machinery, New York, NY, USA, 785–794.

[50] Fanny M Cheung, Fons JR van de Vijver, and Frederick TL Leong. 2011. Toward a new approach to the study of personality in culture. *American Psychologist* 66, 7 (2011), 593.

[51] Gokul Chittaranjan, Jan Blom, and Daniel Gatica-Perez. 2011. Who's who with big-five: Analyzing and classifying personality traits with smartphones. In *Wearable Computers (ISWC), 2011 15th Annual International Symposium on*. IEEE, 29–36.

[52] Gokul Chittaranjan, Jan Blom, and Daniel Gatica-Perez. 2013. Mining large-scale smartphone data for personality studies. *Personal and Ubiquitous Computing* 17, 3 (2013), 433–450.

[53] Hans Christian, Derwin Suhartono, Andry Chowanda, and Kamal Z Zamli. 2021. Text based personality prediction from multiple social media data sources using pre-trained language model and model averaging. *Journal of Big Data* 8, 1 (2021), 68.

[54] Max Coltheart. 1981. The MRC psycholinguistic database. *The Quarterly Journal of Experimental Psychology Section A* 33, 4 (1981), 497–505.

[55] Brian S Connelly and Deniz S Ones. 2010. An other perspective on personality: meta-analytic integration of observers' accuracy and predictive validity. *Psychological bulletin* 136, 6 (2010), 1092.

[56] Stanley Coopersmith. 1959. A method for determining types of self-esteem. *The Journal of Abnormal and Social Psychology* 59, 1 (1959), 87.

[57] Glen Coppersmith, Mark Dredze, and Craig Harman. 2014. Quantifying mental health signals in Twitter. In *Proceedings of the workshop on computational linguistics and clinical psychology: From linguistic signal to clinical reality*. 51–60.

[58] Wrzus Cornelia and Mehl Matthias R. 2015. Lab and/or field? Measuring personality processes and their social consequences. *European Journal of Personality* 29 (2015), 250–271.

[59] C. Cortes and V Vapnik. 1995. Support-Vector Networks. *Machine Learning* 20 (1995).

[60] Paul T Costa and Robert R McCrae. 1992. Normal personality assessment in clinical practice: The NEO Personality Inventory. *Psychological assessment* 4, 1 (1992), 5.

[61] Paul T Costa and Robert R McCrae. 2008. The Revised NEO Personality Inventory (NEO-PI-R). *In G.J. Boyle, G Matthews and D. Saklofske (Eds.). The SAGE handbook of personality theory and assessment* 2 (2008), 179–198.

[62] Marcus Credé, Peter Harms, Sarah Niehorster, and Andrea Gaye-Valentine. 2012. An evaluation of the consequences of using short measures of the Big Five personality traits. *Journal of personality and social psychology* 102, 4 (2012), 874.

[63] Douglas P. Crowne and David Marlowe. 1960. A new scale of social desirability independent of psychopatology. *Journal of Consulting Psychology* 24, 4 (1960), 349–354.

[64] Yves-Alexandre de Montjoye, Jordi Quoidbach, Florent Robic, and Alex Sandy Pentland. 2013. Predicting personality using novel mobile phone-based metrics. In *Social Computing, Behavioral-Cultural Modeling and Prediction*. Springer, 48–55.

[65] Rodrigo de Oliveira, Alexandros Karatzoglou, Pedro Concejero Cerezo, Ana Armenta Lopez de Vicuña, and Nuria Oliver. 2011. Towards a psychographic user model from mobile phone usage. In *CHI'11 Extended Abstracts on Human Factors in Computing Systems*. ACM, 2191–2196.

[66] Jean-Marc Dewaele and Adrian Furnham. 2000. Personality and speech production: A pilot study of second language learners. *Personality and Individual differences* 28, 2 (2000), 355–365.

[67] Sahraoui Dhelim, Nyothiri Aung, Mohammed Amine Bouras, Huansheng Ning, and Erik Cambria. 2022. A survey on personality-aware recommendation systems. *Artificial Intelligence Review* (2022), 1–46.

[68] John M Digman. 1990. Personality structure: Emergence of the five-factor model. *Annual review of psychology* 41, 1 (1990), 417–440.

[69] M Brent Donnellan, Rand D Conger, and Chalandra M Bryant. 2004. The Big Five and enduring marriages. *Journal of Research in personality* 38, 5 (2004), 481–504.

[70] M Brent Donnellan, Frederick L Oswald, Brendan M Baird, and Richard E Lucas. 2006. The mini-IPIP scales: tiny-yet-effective measures of the Big Five factors of personality. *Psychological assessment* 18, 2 (2006), 192.

[71] Kamal El-Demerdash, Reda A El-Khoribi, Mahmoud A Ismail Shoman, and Sherif Abdou. 2022. Deep learning based fusion strategies for personality prediction. *Egyptian Informatics Journal* 23, 1 (2022), 47–53.

[72] Frank Enos, Stefan Benus, Robin L Cautin, Martin Graciarena, Julia Hirschberg, and Elizabeth Shriberg. 2006. Personality factors in human deception detection: comparing human to machine performance. (2006).

[73] Izel Ergu, Zerrin Işık, and İsmail Yankayış. 2019. Predicting personality with twitter data and machine learning models. In *2019 innovations in intelligent systems and applications conference (ASYU)*. IEEE, 1–5.

[74] Adnan Veysel Ertemel and Gökhan Çaylak. 2021. The effect of personality traits on credit score using Myers-Briggs Type Indicator (MBTI) personality types. In *Eurasian Business and Economics Perspectives: Proceedings of the 32nd Eurasia Business and Economics Society Conference*. Springer, 185–199.

[75] Dominique Estival, Tanja Gaustad, Son Bao Pham, Will Radford, and Ben Hutchinson. 2007. Author profiling for English emails. In *Proceedings of the 10th Conference of the Pacific Association for Computational Linguistics (PACLING'07)*. 263–272.

[76] Morgane Evin, Antonio Hidalgo-Munoz, Adolphe James Béquet, Fabien Moreau, Hélène Tattegrain, Catherine Berthelon, Alexandra Fort, and Christophe Jallais. 2022. Personality trait prediction by machine learning using physiological data and driving behavior. *Machine Learning with*

*Applications* 9 (2022), 100353.

[77] Florian Eyben, Martin Wöllmer, and Björn Schuller. 2010. Opensmile: the munich versatile and fast open-source audio feature extractor. In *Proceedings of the 18th ACM international conference on Multimedia*. 1459–1462.

[78] Filipe Falcão, Bárbara Sousa, Daniela SM Pereira, Renato Andrade, Pedro Moreira, Anna Quialheiro, Carlos Jalali, and Patrício Costa. 2023. We vote for the person, not the policies: a systematic review on how personality traits influence voting behaviour. *Discover Psychology* 3, 1 (2023), 1.

[79] Golnoosh Farnadi, Geetha Sitaraman, Mehrdad Rohani, Michal Kosinski, David Stillwell, Marie-Francine Moens, Sergio Davalos, and Martine De Cock. 2014. How are you doing?: emotions and personality in Facebook. In *2nd Workshop on Emotions and Personality in Personalized Services (EMPIRE 2014); workshop at the 22nd Conference on User Modelling, Adaptation and Personalization (UMAP 2014)*. 45–56.

[80] Golnoosh Farnadi, Geetha Sitaraman, Shanu Sushmita, Fabio Celli, Michal Kosinski, David Stillwell, Sergio Davalos, Marie-Francine Moens, and Martine De Cock. 2016. Computational personality recognition in social media. *User modeling and user-adapted interaction* 26 (2016), 109–142.

[81] Golnoosh Farnadi, Shanu Sushmita, Geetha Sitaraman, Nhat Ton, Martine De Cock, and Sergio Davalos. 2014. A multivariate regression approach to personality impression recognition of vloggers. In *Proceedings of the 2014 ACM Multi Media on Workshop on Computational Personality Recognition*. 1–6.

[82] Golnoosh Farnadi, Susana Zoghbi, Marie-Francine Moens, and Martine De Cock. 2013. Recognising personality traits using facebook status updates. In *Proceedings of the international AAAI conference on web and social media*, Vol. 7. 14–18.

[83] L.W. Ferguson. 1952. *Personality Measurement*. McGraw-Hill.

[84] Bruce Ferwerda, Markus Schedl, and Marko Tkalcic. 2015. Predicting personality traits with instagram pictures. In *Proceedings of the 3rd Workshop on Emotions and Personality in Personalized Systems 2015*. 7–10.

[85] Bruce Ferwerda, Markus Schedl, and Marko Tkalcic. 2016. Using instagram picture features to predict users' personality. In *MultiMedia Modeling: 22nd International Conference, MMM 2016, Miami, FL, USA, January 4-6, 2016, Proceedings, Part I 22*. Springer, 850–861.

[86] Bruce Ferwerda and Marko Tkalcic. 2018. Predicting users' personality from Instagram pictures: using visual and/or content features?. In *Proceedings of the 26th conference on user modeling, adaptation and personalization*. 157–161.

[87] Bruce Ferwerda and Marko Tkalcic. 2018. You are what you post: What the content of Instagram pictures tells about users' personality. In *The 23rd International on Intelligent User Interfaces, March 7-11, Tokyo, Japan*. CEUR-WS.

[88] Alan Finder et al. 2006. For some, online persona undermines a resume. *The New York Times* 11 (2006).

[89] William Fleeson. 2001. Toward a structure-and process-integrated view of personality: Traits as density distributions of states. *Journal of personality and social psychology* 80, 6 (2001), 1011.

[90] Tommaso Fornaciari, Fabio Celli, and Massimo Poesio. 2013. The Effect of Personality Type on Deceptive Communication Style. In *Intelligence and Security Informatics Conference (EISIC), 2013 European*. IEEE, 1–6.

[91] Yoav Freund and Robert E. Schapire. 1997. A decision-theoretic generalization of on-line learning and an application to boosting. *J. Comput. System Sci.* 55 (1997), 119–139.

[92] David C. Funder. 2001. Personality. *Annual Review of Psychology* 52, 3 (2001), 197–221.

[93] David C. Funder. 2006. Towards a resolution of the personality triad: Persons, situations, and behaviors. *Journal of Research in Personality* 40 (2006), 21–34.

[94] Christopher M Gallagher, Brent A Stevenor, Andrew Samo, and Samuel T McAbee. 2023. A Short Measure of the Big Five Aspects: Development and Validation of the BFAS-40. *Journal of Personality Assessment* 105, 6 (2023), 719–732.

[95] Peter Zhuowei Gan, Arcot Sowmya, and Gelareh Mohammadi. 2022. Zero-shot Personality Perception From Facial Images. In *Australasian Joint Conference on Artificial Intelligence*. Springer, 43–56.

[96] Adithya V Ganesan, Yash Kumar Lal, August Nilsson, and H Schwartz. 2023. Systematic Evaluation of GPT-3 for Zero-Shot Personality Estimation. In *Proceedings of the 13th Workshop on Computational Approaches to Subjectivity, Sentiment, & Social Media Analysis*. 390–400.

[97] Salvatore Giorgi, João Sedoc, Valentin Barriere, and Shabnam Tafreshi. 2024. Findings of wassa 2024 shared task on empathy and personality detection in interactions. In *Proceedings of the 14th Workshop on Computational Approaches to Subjectivity, Sentiment, & Social Media Analysis*. 369–379.

[98] Matej Gjurković, Vanja Mladen Karan, Iva Vukojević, Mihaela Bošnjak, and Jan Snajder. 2021. PANDORA Talks: Personality and Demographics on Reddit. In *Proceedings of the Ninth International Workshop on Natural Language Processing for Social Media*, Lun-Wei Ku and Cheng-Te Li (Eds.). Association for Computational Linguistics, Online, 138–152. https://doi.org/10.18653/v1/2021.socialnlp-1.12

[99] Joe J. Gladstone, Sandra C. Matz, and Alain Lemaire. 2019. Can Psychological Traits Be Inferred From Spending? Evidence From Transaction Data. *Psychological Science* 30, 7 (2019), 1087–1096.

[100] Jennifer Golbeck. 2016. Predicting personality from social media text. *AIS Transactions on Replication Research* 2, 1 (2016), 2.

[101] Jennifer Golbeck, Cristina Robles, Michon Edmondson, and Karen Turner. 2011. Predicting personality from twitter. In *2011 IEEE third international conference on privacy, security, risk and trust and 2011 IEEE third international conference on social computing*. IEEE, 149–156.

[102] Jennifer Golbeck, Cristina Robles, and Karen Turner. 2011. Predicting personality with social media. In *CHI'11 extended abstracts on human factors in computing systems*. ACM, 253–262.

[103] Lewis R Goldberg. 1982. From Ace to Zombie: Some explorations in the language of personality. *Advances in personality assessment* 1 (1982), 203–234.

[104] Lewis R Goldberg. 1990. An alternative" description of personality": the big-five factor structure. *Journal of personality and social psychology* 59, 6 (1990), 1216.

[105] Leonard V Gordon. 1951. Validities of the forced-choice and questionnaire methods of personality measurement. *Journal of Applied Psychology* 35, 6 (1951), 407.

[106] Louis August Gottschalk and Goldine C Gleser. 1969. *The measurement of psychological states through the content analysis of verbal behavior.* Univ of California Press.

[107] Harrison G Gough. 2012. The California psychological inventory. In *Testing and assessment in counseling practice.* Routledge, 41–65.

[108] David T Graham, John A Stern, and George Winokur. 1958. Experimental investigation of the specificity of attitude hypothesis in psychosomatic disease. *Psychosomatic Medicine* 20, 6 (1958), 446–457.

[109] Sharath Chandra Guntuku, Weisi Lin, Jordan Carpenter, Wee Keong Ng, Lyle H Ungar, and Daniel Preoțiuc-Pietro. 2017. Studying personality through the content of posted and liked images on Twitter. In *Proceedings of the 2017 ACM on web science conference.* 223–227.

[110] Sharath Chandra Guntuku, Lin Qiu, Sujoy Roy, Weisi Lin, and Vinit Jakhetiya. 2015. Do others perceive you as you want them to? Modeling personality based on selfies. In *Proceedings of the 1st international workshop on affect & sentiment in multimedia.* 21–26.

[111] Sharath Chandra Guntuku, Joey Tianyi Zhou, Sujoy Roy, Weisi Lin, and Ivor W Tsang. 2016. Who likes what and, why?'insights into modeling users' personality based on image 'likes. *IEEE Transactions on Affective Computing* 9, 1 (2016), 130–143.

[112] Gareth Hagger-Johnson, Vincent Egan, and David Stillwell. 2011. Are social networking profiles reliable indicators of sensational interests? *Journal of Research in Personality* 45, 1 (2011), 71–76.

[113] Pietter Haizel, Grace Vernanda, Keyzia Alexandra Wawolangi, and Novita Hanafiah. 2021. Personality assessment video game based on the five-factor model. *Procedia Computer Science* 179 (2021), 566–573.

[114] Noptovius Halimawan, Derwin Suhartono, Aryo Pradipta Gema, and Rezki Yunanda. 2022. BERT and ULMFiT Ensemble for Personality Prediction from Indonesian Social Media Text. In *2022 International Symposium on Information Technology and Digital Innovation (ISITDI).* IEEE, 156–161.

[115] Yuval Noah Harari. 2024. *Nexus: A brief history of information networks from the stone age to AI.* Signal.

[116] James V Haxby, Elizabeth A Hoffman, and M Ida Gobbini. 2000. The distributed human neural system for face perception. *Trends in cognitive sciences* 4, 6 (2000), 223–233.

[117] Christina U Heinrich and Peter Borkenau. 1998. Deception and deception detection: The role of cross-modal Inconsistency. *Journal of Personality* 66, 5 (1998), 687–712.

[118] Joanne Hinds and Adam N Joinson. 2024. Digital data and personality: A systematic review and meta-analysis of human perception and computer prediction. *Psychological bulletin* 150, 6 (2024), 727.

[119] Robert Hogan and Joyce Hogan. 1995. Hogan personality inventory. *Journal of Applied Psychology* (1995).

[120] Linmei Hu, Hongyu He, Duokang Wang, Ziwang Zhao, Yingxia Shao, and Liqiang Nie. 2024. LLM vs Small Model? Large Language Model Based Text Augmentation Enhanced Personality Detection Model. In *Proceedings of the AAAI Conference on Artificial Intelligence*, Vol. 38. 18234–18242.

[121] Rong Hu and Pearl Pu. 2009. Acceptance issues of personality-based recommender systems. In *Proceedings of the third ACM conference on Recommender systems.* 221–224.

[122] Rong Hu and Pearl Pu. 2010. Using personality information in collaborative filtering for new users. *Recommender Systems and the Social Web* 17 (2010), 60–70.

[123] Rong Hu and Pearl Pu. 2011. Enhancing collaborative filtering systems with personality information. In *Proceedings of the fifth ACM conference on Recommender systems.* 197–204.

[124] Jen-tse Huang, Wenxuan Wang, Eric John Li, Man Ho Lam, Shujie Ren, Youliang Yuan, Wenxiang Jiao, Zhaopeng Tu, and Michael R. Lyu. 2024. On the Humanity of Conversational AI: Evaluating the Psychological Portrayal of LLMs. In *The Twelfth International Conference on Learning Representations.* OpenReview.net.

[125] Jen-tse Huang, Wenxiang Jiao, Man Ho Lam, Eric John Li, Wenxuan Wang, and Michael Lyu. 2024. On the Reliability of Psychological Scales on Large Language Models. In *Proceedings of the 2024 Conference on Empirical Methods in Natural Language Processing*, Yaser Al-Onaizan, Mohit Bansal, and Yun-Nung Chen (Eds.). Association for Computational Linguistics, 6152–6173.

[126] Francisco Iacobelli and Aron Culotta. 2013. Too neurotic, not too friendly: structured personality classification on textual data. In *Proceedings of the International AAAI Conference on Web and Social Media*, Vol. 7. 19–22.

[127] Francisco Iacobelli, Alastair J Gill, Scott Nowson, and Jon Oberlander. 2011. Large scale personality classification of bloggers. In *Affective Computing and Intelligent Interaction.* Springer, 568–577.

[128] Alexei V Ivanov, Giuseppe Riccardi, Adam J Sporka, and Jakub Franc. 2011. Recognition of Personality Traits from Human Spoken Conversations.. In *Interspeech.* 1549–1552.

[129] Chris J. Jackson and Leslie J. Francis. 1998. Interpreting the correlation between neuroticism and lie scale scores. *Personality and Individual Differences* 26, 1 (1998), 59–63.

[130] Julio C. S. Jacques Junior, Yağmur Güçlütürk, Marc Pérez, Umut Güçlü, Carlos Andujar, Xavier Baró, Hugo Jair Escalante, Isabelle Guyon, Marcel A. J. van Gerven, Rob van Lier, and Sergio Escalera. 2022. First Impressions: A Survey on Vision-Based Apparent Personality Trait Analysis. *IEEE Transactions on Affective Computing* 13, 1 (2022), 75–95.

[131] Nahariah Jaffar, Hasnah Haron, Takiah Mohd Iskandar, and Arfah Salleh. 2011. Fraud risk assessment and detection of fraud: the moderating effect of personality. *International Journal of Business and Management* 6, 7 (2011), 40.

[132] Madhura Jayaratne and Buddhi Jayatilleke. 2020. Predicting personality using answers to open-ended interview questions. *IEEE Access* 8 (2020), 115345–115355.

[133] Guangyuan Jiang, Manjie Xu, Song-Chun Zhu, Wenjuan Han, Chi Zhang, and Yixin Zhu. 2023. Evaluating and inducing personality in pre-trained language models. In *Proceedings of the 37th International Conference on Neural Information Processing Systems* (New Orleans, LA, USA) *(NIPS '23)*. Curran Associates Inc., Red Hook, NY, USA, Article 466, 22 pages.

[134] Hang Jiang, Xiajie Zhang, Xubo Cao, Cynthia Breazeal, Deb Roy, and Jad Kabbara. 2024. PersonaLLM: Investigating the Ability of Large Language Models to Express Personality Traits. In *Findings of the Association for Computational Linguistics: NAACL 2024*, Kevin Duh, Helena Gomez, and Steven Bethard (Eds.). Association for Computational Linguistics, Mexico City, Mexico, 3605–3627.

[135] Oliver P John. 1990. The" Big Five" factor taxonomy: Dimensions of personality in the natural language and in questionnaires. *Handbook of personality: Theory and research* (1990).

[136] Oliver P John, Richard W Robins, and Lawrence A Pervin. 2010. *Handbook of personality: Theory and research*. Guilford Press.

[137] John A Johnson. 2014. Measuring thirty facets of the Five Factor Model with a 120-item public domain inventory: Development of the IPIP-NEO-120. *Journal of research in personality* 51 (2014), 78–89.

[138] Daniel N. Jones and Delroy L. Paulhus. 2014. Introducing the Short Dark Triad (SD3): A Brief Measure of Dark Personality Traits. *Assessment* 21, 1 (2014), 28–41.

[139] Carl G Jung. 1921. Psychologische Typen, Duesseldorf.

[140] Alexander Kachur, Evgeny Osin, Denis Davydov, Konstantin Shutilov, and Alexey Novokshonov. 2020. Assessing the Big Five personality traits using real-life static facial images. *Scientific Reports* 10, 1 (2020), 8487.

[141] Kyriaki Kalimeri, Bruno Lepri, and Fabio Pianesi. 2010. Causal-modelling of personality traits: extraversion and locus of control. In *Proceedings of the 2nd international workshop on Social signal processing*. 41–46.

[142] Kyriaki Kalimeri, Bruno Lepri, and Fabio Pianesi. 2013. Going beyond traits: Multimodal classification of personality states in the wild. In *Proceedings of the 15th ACM on International conference on multimodal interaction*. ACM, 27–34.

[143] Onno Kampman, Elham J Barezi, Dario Bertero, and Pascale Fung. 2018. Investigating audio, visual, and text fusion methods for end-to-end automatic personality prediction. *arXiv preprint arXiv:1805.00705* (2018).

[144] Arvid Karsvall. 2002. Personality preferences in graphical interface design. In *Proceedings of the second Nordic conference on Human-computer interaction*. 217–218.

[145] KMGS Karunarathna, MPRIR Silva, and RAHM Rupasingha. 2023. Ensemble Learning Approach for Identifying Personality Traits based on Individuals' Behavior. *EDITORIAL NOTE* (2023), 107.

[146] Vishal Kaushal and Manasi Patwardhan. 2018. Emerging trends in personality identification using online social networks—a literature survey. *ACM Transactions on Knowledge Discovery from Data (TKDD)* 12, 2 (2018), 1–30.

[147] Amirmohammad Kazameini, Samin Fatehi, Yash Mehta, Sauleh Eetemadi, and Erik Cambria. 2020. Personality trait detection using bagged svm over bert word embedding ensembles. *arXiv preprint arXiv:2010.01309* (2020).

[148] Edbert Ivan Sebastian Kelvin and Suhartono Derwin. 2023. Utilizing IndoBERT in Predicting Personality from Twitter Posts Using Bahasa Indonesia. *ICIC Express Letters* 17, 01 (2023), 123.

[149] Maurice Kendall and Jean D. Gibbons. 1990. *Rank correlation methods*. New York: Oxford University Press.

[150] Katia Lida Kermanidis. 2012. Mining authors' personality traits from modern greek spontaneous text. In *Proc. of Workshop on Corpora for Research on Emotion Sentiment & Social Signals, in conjunction with LREC*. Citeseer, 90–93.

[151] Margaret L. Kern, Paul X. McCarthy, Deepanjan Chakrabarty, and Marian-Andrei Rizoiu. 2019. Social media-predicted personality traits and values can help match people to their ideal jobs. *Proceedings of the National Academy of Sciences* 16 (2019), 26459–26464.

[152] Lawrence J Klinkert, Steph Buongiorno, and Corey Clark. 2024. Evaluating the Efficacy of LLMs to Emulate Realistic Human Personalities. In *Proceedings of the AAAI Conference on Artificial Intelligence and Interactive Digital Entertainment*, Vol. 20. 65–75.

[153] Michal Kosinski, Yoram Bachrach, Pushmeet Kohli, David Stillwell, and Thore Graepel. 2014. Manifestations of user personality in website choice and behaviour on online social networks. *Machine learning* 95, 3 (2014), 357–380.

[154] Michal Kosinski, Sandra C Matz, Samuel D Gosling, Vesselin Popov, and David Stillwell. 2015. Facebook as a research tool for the social sciences: Opportunities, challenges, ethical considerations, and practical guidelines. *American psychologist* 70, 6 (2015), 543.

[155] Michal Kosinski, David Stillwell, and Thore Graepel. 2013. Private traits and attributes are predictable from digital records of human behavior. *Proceedings of the national academy of sciences* 110, 15 (2013), 5802–5805.

[156] Ivar Krumpal. 2013. Determinants of social desirability bias in sensitive surveys: a literature review. *Quality & Quantity* 47, 4 (2013), 2025–2047.

[157] Akshi Kumar, Rohit Beniwal, and Dipika Jain. 2023. Personality detection using kernel-based ensemble model for leveraging social psychology in online networks. *ACM Transactions on Asian and Low-Resource Language Information Processing* 22, 5 (2023), 1–20.

[158] Aditi V Kunte and Suja Panicker. 2019. Using textual data for personality prediction: a machine learning approach. In *2019 4th international conference on information systems and computer networks (ISCON)*. IEEE, 529–533.

[159] Vasileios Lampos, Nikolaos Aletras, Daniel Preoţiuc-Pietro, and Trevor Cohn. 2014. Predicting and characterising user impact on Twitter. In *Proceedings of the 14th Conference of the European Chapter of the Association for Computational Linguistics*. 405–413.

[160] Kibeom Lee and Michael C Ashton. 2018. Psychometric properties of the HEXACO-100. *Assessment* 25, 5 (2018), 543–556.

[161] Bruno Lepri, Nadia Mana, Alessandro Cappelletti, Fabio Pianesi, and Massimo Zancanaro. 2009. Modeling the personality of participants during group interactions. In *User Modeling, Adaptation, and Personalization: 17th International Conference, UMAP 2009, formerly UM and AH, Trento, Italy, June 22-26, 2009. Proceedings 17*. Springer, 114–125.

[162] Bruno Lepri, Ramanathan Subramanian, Kyriaki Kalimeri, Jacopo Staiano, Fabio Pianesi, and Nicu Sebe. 2010. Employing social gaze and speaking activity for automatic determination of the extraversion trait. In *International Conference on Multimodal Interfaces and the Workshop on Machine Learning for Multimodal Interaction*. ACM, 7.

[163] Bruno Lepri, Ramanathan Subramanian, Kyriaki Kalimeri, Jacopo Staiano, Fabio Pianesi, and Nicu Sebe. 2012. Connecting meeting behavior with extraversion—A systematic study. *IEEE Transactions on Affective Computing* 3, 4 (2012), 443–455.

[164] Wenyu Li, Xin Hu, Xuefei Long, Lilu Tang, Jingjing Chen, Fei Wang, and Dan Zhang. 2020. EEG responses to emotional videos can quantitatively predict big-five personality traits. *Neurocomputing* 415 (2020), 368–381.

[165] Tze Wei Liew and Su-Mae Tan. 2016. Virtual agents with personality: Adaptation of learner-agent personality in a virtual learning environment. In *2016 Eleventh International Conference on Digital Information Management (ICDIM)*. IEEE, 157–162.

[166] Shi Min Lim, Chyi Wey Claudine Shiau, Ling Jie Cheng, and Ying Lau. 2022. Chatbot-delivered psychotherapy for adults with depressive and anxiety symptoms: a systematic review and meta-regression. *Behavior Therapy* 53, 2 (2022), 334–347.

[167] Leqi Liu, Daniel Preotiuc-Pietro, Zahra Riahi Samani, Mohsen E Moghaddam, and Lyle Ungar. 2016. Analyzing personality through social media profile picture choice. In *Tenth international AAAI conference on web and social media*.

[168] PH Lodhi, Savita Deo, and Vivek M Belhekar. 2002. The five-factor model of personality: Measurement and correlates in the Indian context. *The five-factor model of personality across cultures* (2002), 227–248.

[169] David Lowe. 2004. Distinctive image features from scale-invariant keypoints. *International Journal of Computer Vision* 60, 2 (2004).

[170] Kim Luyckx and Walter Daelemans. 2008. Personae: a Corpus for Author and Personality Prediction from Text.. In *LREC*.

[171] François Mairesse and Marilyn Walker. 2006. Automatic recognition of personality in conversation. In *Proceedings of the Human Language Technology Conference of the NAACL, Companion Volume: Short Papers*. Association for Computational Linguistics, 85–88.

[172] François Mairesse and Marilyn Walker. 2007. PERSONAGE: Personality generation for dialogue. In *Proceedings of the 45th annual meeting of the association of computational linguistics*. 496–503.

[173] François Mairesse, Marilyn Walker, et al. 2006. Words mark the nerds: Computational models of personality recognition through language. In *Proceedings of the annual meeting of the cognitive science society*, Vol. 28.

[174] François Mairesse, Marilyn A Walker, Matthias R Mehl, and Roger K Moore. 2007. Using linguistic cues for the automatic recognition of personality in conversation and text. *Journal of artificial intelligence research* 30 (2007), 457–500.

[175] Patrick M Markey and Charlotte N Markey. 2009. A brief assessment of the interpersonal circumplex: The IPIP-IPC. *Assessment* 16, 4 (2009), 352–361.

[176] Dejan Markovikj, Sonja Gievska, Michal Kosinski, and David Stillwell. 2013. Mining facebook data for predictive personality modeling. 7, 2 (2013), 23–26.

[177] Sandra C Matz and Gabriella M Harari. 2021. Personality–place transactions: Mapping the relationships between Big Five personality traits, states, and daily places. *Journal of Personality and Social Psychology* 120, 5 (2021), 1367.

[178] Robert R McCrae. 1989. Why I advocate the five-factor model: Joint factor analyses of the NEO-PI with other instruments. *Personality psychology: Recent trends and emerging directions* (1989), 237–245.

[179] Robert R McCrae and Paul T Costa. 2003. *Personality in adulthood: A five-factor theory perspective*. Guilford Press.

[180] Robert R McCrae and Oliver P John. 1992. An introduction to the five-factor model and its applications. *Journal of personality* 60, 2 (1992), 175–215.

[181] Marshall McLuhan. 2017. The medium is the message. In *Communication theory*. Routledge, 390–402.

[182] Matthias R Mehl, Samuel D Gosling, and James W Pennebaker. 2006. Personality in its natural habitat: manifestations and implicit folk theories of personality in daily life. *Journal of personality and social psychology* 90, 5 (2006), 862.

[183] Yash Mehta, Samin Fatehi, Amirmohammad Kazameini, Clemens Stachl, Erik Cambria, and Sauleh Eetemadi. 2020. Bottom-up and top-down: Predicting personality with psycholinguistic and language model features. In *2020 IEEE International Conference on Data Mining (ICDM)*. IEEE, 1184–1189.

[184] Yash Mehta, Navonil Majumder, Alexander Gelbukh, and Erik Cambria. 2020. Recent trends in deep learning based personality detection. *Artificial Intelligence Review* 53 (2020), 2313–2339.

[185] Alessandro B Melchiorre and Markus Schedl. 2020. Personality correlates of music audio preferences for modelling music listeners. In *Proceedings of the 28th ACM conference on user modeling, adaptation and personalization*. 313–317.

[186] Atsunori Minamikawa and Hiroyuki Yokoyama. 2011. Blog tells what kind of personality you have: egogram estimation from Japanese weblog. In *Proceedings of the ACM 2011 conference on Computer supported cooperative work*. 217–220.

[187] Li Mo, Xiaosan Zhang, Yabin Lin, Zhenghui Yuan, and Zengjun Peng. 2023. Consumers' attitudes towards online advertising: A model of personalization, informativeness, privacy concern and flow experience. *Sustainability* 15, 5 (2023), 4090.

[188] Saif Mohammad and Svetlana Kiritchenko. 2013. Using nuances of emotion to identify personality. In *Proceedings of the International AAAI Conference on Web and Social Media*, Vol. 7. 27–30.

[189] Gelareh Mohammadi, Antonio Origlia, Maurizio Filippone, and Alessandro Vinciarelli. 2012. From speech to personality: Mapping voice quality and intonation into personality differences. In *Proceedings of the 20th ACM international conference on Multimedia*. 789–792.

[190] Gelareh Mohammadi and Alessandro Vinciarelli. 2012. Automatic personality perception: Prediction of trait attribution based on prosodic features. *Affective Computing, IEEE Transactions on* 3, 3 (2012), 273–284.

[191] Gelareh Mohammadi, Alessandro Vinciarelli, and Marcello Mortillaro. 2010. The voice of personality: Mapping nonverbal vocal behavior into trait attributions. In *Proceedings of the 2nd international workshop on Social signal processing*. ACM, 17–20.

[192] Sara Zannone Gabriella M Harari A Aldo Faisal Aleksandar Matic Mohammed Khwaja, Sumer S Vaid. 2019. Modeling personality vs. modeling personalidad: In-the-wild mobile data analysis in five countries suggests cultural impact on personality models. *Proceedings of the ACM on Interactive, Mobile, Wearable and Ubiquitous Technologies* 3 (2019), 1–24. Issue 3.

[193] Isabel Briggs Myers et al. 1962. *The Myers-Briggs type indicator*. Vol. 34. Consulting Psychologists Press Palo Alto, CA.

[194] Ghinaa Zain Nabiilah and Derwin Suhartono. 2023. Personality Classification Based on Textual Data using Indonesian Pre-Trained Language Model and Ensemble Majority Voting. *Revue d'Intelligence Artificielle* 37, 1 (2023).

[195] Yair Neuman and Yochai Cohen. 2024. A Data Set of Synthetic Utterances for Computational Personality Analysis. *Scientific data* 11, 1 (2024), 623.

[196] Thin Nguyen, Dinh Phung, Brett Adams, and Svetha Venkatesh. 2011. Towards discovery of influence and personality traits through social link prediction. In *Proceedings of the International AAAI Conference on Web and Social Media*, Vol. 5. 566–569.

[197] Dong Nie, Zengda Guan, Bibo Hao, Shuotian Bai, and Tingshao Zhu. 2014. Predicting personality on social media with semi-supervised learning. In *2014 IEEE/WIC/ACM International Joint Conferences on Web Intelligence (WI) and Intelligent Agent Technologies (IAT)*, Vol. 2. IEEE, 158–165.

[198] Warren T Norman. 1963. Personality measurement, faking, and detection: An assessment method for use in personnel selection. *Journal of Applied Psychology* 47, 4 (1963), 225.

[199] Warren T Norman. 1963. Toward an adequate taxonomy of personality attributes: Replicated factor structure in peer nomination personality ratings. *The journal of abnormal and social psychology* 66, 6 (1963), 574.

[200] Oded Nov and Chen Ye. 2008. Personality and technology acceptance: Personal innovativeness in IT, openness and resistance to change. In *Proceedings of the 41st annual Hawaii international conference on system sciences (HICSS 2008)*. IEEE, 448–448.

[201] Sam Nunn. 2005. Preventing the next terrorist attack: The theory and practice of homeland security information systems. *Journal of Homeland Security and Emergency Management* 2, 3 (2005).

[202] Jon Oberlander and Alastair J Gill. 2006. Language with character: A stratified corpus comparison of individual differences in e-mail communication. *Discourse processes* 42, 3 (2006), 239–270.

[203] Jon Oberlander and Scott Nowson. 2006. Whose thumb is it anyway?: classifying author personality from weblog text. In *Proceedings of the COLING/ACL on Main conference poster sessions*. Association for Computational Linguistics, 627–634.

[204] Veronica Ong, Anneke DS Rahmanto, Derwin Suhartono, Aryo E Nugroho, Esther W Andangsari, Muhamad N Suprayogi, et al. 2017. Personality prediction based on Twitter information in Bahasa Indonesia. In *2017 federated conference on computer science and information systems (FedCSIS)*. IEEE, 367–372.

[205] Alvaro Ortigosa, Rosa M Carro, and José Ignacio Quiroga. 2014. Predicting user personality by mining social interactions in Facebook. *Journal of computer and System Sciences* 80, 1 (2014), 57–71.

[206] Julie E Owen, Duhita Mahatmya, and Rebecca Carter. 2020. Dominance, influence, steadiness, and conscientiousness (DISC) assessment tool. In *Encyclopedia of personality and individual differences*. Springer, 1186–1189.

[207] Maxwell Jerome Papurt. 1930. A study of the Woodworth Psychoneurotic Inventory with suggested revision. *The Journal of Abnormal and Social Psychology* 25, 3 (1930), 335.

[208] Gregory Park, H Andrew Schwartz, Johannes C Eichstaedt, Margaret L Kern, Michal Kosinski, David J Stillwell, Lyle H Ungar, and Martin EP Seligman. 2015. Automatic personality assessment through social media language. *Journal of personality and social psychology* 108, 6 (2015), 934.

[209] Frank T Passini and Warren T Norman. 1966. A universal conception of personality structure? *Journal of personality and social psychology* 4, 1 (1966), 44.

[210] Dean Peabody and Boele De Raad. 2002. The substantive nature of psycholexical personality factors: A comparison across languages. *Journal of personality and Social Psychology* 83, 4 (2002), 983.

[211] Dean Peabody and Lewis R Goldberg. 1989. Some determinants of factor structures from personality-trait descriptors. *Journal of personality and social psychology* 57, 3 (1989), 552.

[212] James W Pennebaker, Martha E Francis, and Roger J Booth. 2001. Linguistic inquiry and word count: LIWC 2001. *Mahway: Lawrence Erlbaum Associates* 71, 2001 (2001), 2001.

[213] James W Pennebaker and Laura A King. 1999. Linguistic styles: language use as an individual difference. *Journal of personality and social psychology* 77, 6 (1999), 1296.

[214] Heinrich Peters and Sandra C Matz. 2024. Large language models can infer psychological dispositions of social media users. *PNAS nexus* 3, 6 (2024), pgae231.

[215] Matthew E. Peters, Mark Neumann, Mohit Iyyer, Matt Gardner, Christopher Clark, Kenton Lee, and Luke Zettlemoyer. 2018. Deep Contextualized Word Representations. In *Proc. of the 2018 Conference of the North American Chapter of the Association for Computational Linguistics*. 2227–2237.

[216] James G Phillips, Sarah Butt, and Alex Blaszczynski. 2006. Personality and self-reported use of mobile phones for games. *CyberPsychology & Behavior* 9, 6 (2006), 753–758.

[217] Fabio Pianesi, Nadia Mana, Alessandro Cappelletti, Bruno Lepri, and Massimo Zancanaro. 2008. Multimodal recognition of personality traits in social interactions. In *Proceedings of the 10th international conference on Multimodal interfaces*. 53–60.

[218] Fabio Pianesi, Massimo Zancanaro, Bruno Lepri, and Alessandro Cappelletti. 2007. A multimodal annotated corpus of consensus decision making meetings. *Language Resources and Evaluation* 41 (2007), 409–429.

[219] Barbara Plank and Dirk Hovy. 2015. Personality Traits on Twitter - or - How to Get 1,500 Personality Tests in a Week. *6TH Workshop on computational approaches to subjectivity, sentiment and social media analysis WASSA 2015* (2015), 92.

[220] Tim Polzehl, Sebastian Möller, and Florian Metze. 2010. Automatically assessing personality from speech. In *2010 IEEE Fourth International Conference on Semantic Computing*. IEEE, 134–140.

[221] Soujanya Poria, Alexandar Gelbukh, Basant Agarwal, Erik Cambria, and Newton Howard. 2013. Common sense knowledge based personality recognition from text. In *Advances in Soft Computing and Its Applications: 12th Mexican International Conference on Artificial Intelligence, MICAI 2013, Mexico City, Mexico, November 24-30, 2013, Proceedings, Part II 12*. Springer, 484–496.

[222] Daniele Quercia, Michal Kosinski, David Stillwell, and Jon Crowcroft. 2011. Our twitter profiles, our selves: Predicting personality with twitter. In *2011 IEEE third international conference on privacy, security, risk and trust and 2011 IEEE third international conference on social computing*. IEEE, 180–185.

[223] Daniele Quercia, Renaud Lambiotte, David Stillwell, Michal Kosinski, and Jon Crowcroft. 2012. The personality of popular facebook users. In *Proceedings of the ACM 2012 conference on computer supported cooperative work*. 955–964.

[224] Muhannad Quwaider, Abdullah Alabed, and Rehab Duwairi. 2023. Shooter video games for personality prediction using five factor model traits and machine learning. *Simulation Modelling Practice and Theory* 122 (2023), 102665.

[225] Colin Raffel, Noam Shazeer, Adam Roberts, Katherine Lee, Sharan Narang, Michael Matena, Yanqi Zhou, Wei Li, and Peter J. Liu. 2020. Exploring the limits of transfer learning with a unified text-to-text transformer. *Journal of Machine Learning Research* 21 (2020), 1–67.

[226] Majid Ramezani, Mohammad-Reza Feizi-Derakhshi, and Mohammad-Ali Balafar. 2022. Knowledge Graph-Enabled Text-Based Automatic Personality Prediction. *Computational intelligence and neuroscience* 2022, 1 (2022), 3732351.

[227] Majid Ramezani, Mohammad-Reza Feizi-Derakhshi, and Mohammad-Ali Balafar. 2022. Text-based automatic personality prediction using KGrAt-Net: a knowledge graph attention network classifier. *Scientific Reports* 12, 1 (2022), 21453.

[228] Majid Ramezani, Mohammad-Reza Feizi-Derakhshi, Mohammad-Ali Balafar, Meysam Asgari-Chenaghlu, Ali-Reza Feizi-Derakhshi, Narjes Nikzad-Khasmakhi, Mehrdad Ranjbar-Khadivi, Zoleikha Jahanbakhsh-Nagadeh, Elnaz Zafarani-Moattar, and Taymaz Akan. 2022. Automatic personality prediction: an enhanced method using ensemble modeling. *Neural Computing and Applications* 34, 21 (2022), 18369–18389.

[229] Francisco Manuel Rangel Pardo, Fabio Celli, Paolo Rosso, Martin Potthast, Benno Stein, and Walter Daelemans. 2015. Overview of the 3rd Author Profiling Task at PAN 2015. In *CLEF 2015 evaluation labs and workshop working notes papers*. 1–8.

[230] Brent W. Roberts and Hee J. Yoon. 2022. Personality Psychology. *Annual Review of Psychology* 73 (2022), 489–516.

[231] William Peter Robinson and Howard Giles. 1990. *Handbook of language and social psychology*. Wiley Chichester, UK.

[232] GV Rohit, K Rakesh Bharadwaj, R Hemanth, Bariti Pruthvi, and MV Manoj Kumar. 2020. Machine intelligence based personality prediction using social profile data. In *2020 Third International Conference on Smart Systems and Inventive Technology (ICSSIT)*. IEEE, 1003–1008.

[233] Alexandra Roshchina, John Cardiff, and Paolo Rosso. 2011. A comparative evaluation of personality estimation algorithms for the twin recommender system. In *Proceedings of the 3rd international workshop on Search and mining user-generated contents*. 11–18.

[234] Ethan Rublee, Vincent Rabaud, Kurt Konolige, and Gary Bradski. 2011. ORB: an efficient alternative to SIFT or SURF. In *Computer Vision (ICCV), 2011 IEEE International Conference on*. IEEE, 2564–2571.

[235] Gregorius Ryan, Pricillia Katarina, and Derwin Suhartono. 2023. Mbti personality prediction using machine learning and smote for balancing data based on statement sentences. *Information* 14, 4 (2023), 217.

[236] Jitao Sang, Huaiwen Zhang, and Changsheng Xu. 2016. Visual BFI: An exploratory study for image-based personality test. In *Advances in Multimedia Information Processing-PCM 2016: 17th Pacific-Rim Conference on Multimedia, Xi' an, China, September 15-16, 2016, Proceedings, Part I*. Springer, 95–106.

[237] Olga C Santos. 2016. Emotions and personality in adaptive e-learning systems: an affective computing perspective. *Emotions and personality in personalized services: Models, evaluation and applications* (2016), 263–285.

[238] Chandrima Sarkar, Sumit Bhatia, Arvind Agarwal, and Juan Li. 2014. Feature analysis for computational personality recognition using youtube personality data set. In *Proceedings of the 2014 ACM multi media on workshop on computational personality recognition*. 11–14.

[239] G Nave DJ Stillwell SC Matz, M Kosinski. 2017. Psychological targeting as an effective approach to digital mass persuasion. *Proceedings of the National Academy of Sciences* 114 (2017), 12714–12719. Issue 48.

[240] SS Vaid H Peters GM Harari M Cerf SC Matz, JD Teeny. 2024. The potential of generative AI for personalized persuasion at scale. *Scientific Reports* 14 (2024), 4692. Issue 1.

[241] Björn Schuller, Stefan Steidl, Anton Batliner, Elmar Nöth, Alessandro Vinciarelli, Felix Burkhardt, Rob Van Son, Felix Weninger, Florian Eyben, Tobias Bocklet, et al. 2012. The interspeech 2012 speaker trait challenge. In *INTERSPEECH 2012, Portland, OR, USA*.

[242] Andrew H Schwartz, Johannes C Eichstaedt, Margaret L Kern, Lukasz Dziurzynski, Stephanie M Ramones, Megha Agrawal, Achal Shah, Michal Kosinski, David Stillwell, Martin EP Seligman, et al. 2013. Personality, Gender, and Age in the Language of Social Media: The Open-Vocabulary Approach. *PloS one* 8, 9 (2013), 773–791.

[243] Cristina Segalin, Fabio Celli, Luca Polonio, Michal Kosinski, David Stillwell, Nicu Sebe, Marco Cristani, and Bruno Lepri. 2017. What your Facebook profile picture reveals about your personality. In *Proceedings of the 25th ACM international conference on Multimedia*. 460–468.

[244] Cristina Segalin, Dong Seon Cheng, and Marco Cristani. 2017. Social profiling through image understanding: Personality inference using convolutional neural networks. *Computer Vision and Image Understanding* 156 (2017), 34–50.

[245] Cristina Segalin, Alessandro Perina, Marco Cristani, and Alessandro Vinciarelli. 2016. The pictures we like are our image: continuous mapping of favorite pictures into self-assessed and attributed personality traits. *IEEE Transactions on Affective Computing* 8, 2 (2016), 268–285.

[246] Doaa Shawky and Ashraf Badawi. 2019. Towards a personalized learning experience using reinforcement learning. *Machine learning paradigms: Theory and application* (2019), 169–187.

[247] Siyang Song, Shashank Jaiswal, Enrique Sanchez, Georgios Tzimiropoulos, Linlin Shen, and Michel Valstar. 2021. Self-supervised learning of person-specific facial dynamics for automatic personality recognition. *IEEE Transactions on Affective Computing* 14, 1 (2021), 178–195.

[248] Ruchir Srivastava, Jiashi Feng, Sujoy Roy, Shuicheng Yan, and Terence Sim. 2012. Don't ask me what I'm like, just watch and listen. In *Proceedings of the 20th ACM international conference on Multimedia*. 329–338.

[249] Jacopo Staiano, Bruno Lepri, Nadav Aharony, Fabio Pianesi, Nicu Sebe, and Alex Pentland. 2012. Friends don't lie: inferring personality traits from social network structure. In *Proceedings of the 2012 ACM conference on ubiquitous computing*. ACM, 321–330.

[250] Jacopo Staiano, Bruno Lepri, Ramanathan Subramanian, Nicu Sebe, and Fabio Pianesi. 2011. Automatic modeling of personality states in small group interactions. In *Proceedings of the 19th ACM international conference on Multimedia*. ACM, 989–992.

[251] Ramanathan Subramanian, Yan Yan, Jacopo Staiano, Oswald Lanz, and Nicu Sebe. 2013. On the relationship between head pose, social attention and personality prediction for unstructured and dynamic group interactions. In *Proceedings of the 15th ACM on International conference on multimodal interaction*. 3–10.

[252] Chanchal Suman, Sriparna Saha, Aditya Gupta, Saurabh Kumar Pandey, and Pushpak Bhattacharyya. 2022. A multi-modal personality prediction system. *Knowledge-Based Systems* 236 (2022), 107715.

[253] Chris Sumner, Alison Byers, Rachel Boochever, and Gregory J Park. 2012. Predicting dark triad personality traits from twitter usage and a linguistic analysis of tweets. In *2012 11th international conference on machine learning and applications*, Vol. 2. IEEE, 386–393.

[254] Shlomo Argamon Shlomo, Shlomo Argamon, Sushant Dhawle, and James W. Pennebaker. 2005. Lexical Predictors Of Personality Type. In *In Proceedings of the Joint Annual Meeting of the Interface and the Classification Society of North America*. 1–16.

[255] Ludovic Terren Ludovic Terren and Rosa Borge-Bravo Rosa Borge-Bravo. 2021. Echo chambers on social media: A systematic review of the literature. *Review of Communication Research* 9 (2021).

[256] Menasha Thilakaratne, Ruvan Weerasinghe, and Sujan Perera. 2016. Knowledge-driven approach to predict personality traits by leveraging social media data. In *2016 IEEE/WIC/ACM International Conference on Web Intelligence (WI)*. IEEE, 288–295.

[257] Marko Tkalcic, Matevz Kunaver, Andrej Košir, and Jurij Tasic. 2011. Addressing the new user problem with a personality based user similarity measure. In *First International Workshop on Decision Making and Recommendation Acceptance Issues in Recommender Systems (DEMRA 2011)*, Vol. 740. 106–111.

[258] Marc Tomlinson, David Hinote, and David Bracewell. 2013. Predicting conscientiousness through semantic analysis of facebook posts. In *Proceedings of the International AAAI Conference on Web and Social Media*, Vol. 7. 31–34.

[259] Natkamon Tovanich, Simone Centellegher, Nacéra Bennacer Seghouani, Joe Gladstone, Sandra Matz, and Bruno Lepri. 2021. Inferring psychological traits from spending categories and dynamic consumption patterns. *EPJ Data Science* 10, 1 (2021), 24.

[260] Timothy J Trull and David C Geary. 1997. Comparison of the big-five factor structure across samples of Chinese and American adults. *Journal of Personality Assessment* 69, 2 (1997), 324–341.

[261] Simon Tucker and Steve Whittaker. 2005. Accessing multimodal meeting data: Systems, problems and possibilities. In *Machine Learning for Multimodal Interaction: First International Workshop, MLMI 2004, Martigny, Switzerland, June 21-23, 2004, Revised Selected Papers 1*. Springer, 1–11.

[262] James S Uleman, S Adil Saribay, and Celia M Gonzalez. 2008. Spontaneous inferences, implicit impressions, and implicit theories. *Annu. Rev. Psychol.* 59 (2008), 329–360.

[263] Carles Ventura, David Masip, and Agata Lapedriza. 2017. Interpreting cnn models for apparent personality trait regression. In *Proceedings of the IEEE conference on computer vision and pattern recognition workshops*. 55–63.

[264] Ben Verhoeven, Walter Daelemans, and Tom De Smedt. 2013. Ensemble methods for personality recognition. In *Proceedings of the International AAAI Conference on Web and Social Media*, Vol. 7. 35–38.

[265] Alessandro Vinciarelli and Gelareh Mohammadi. 2014. A survey of personality computing. *IEEE Transactions on Affective Computing* 5, 3 (2014), 273–291.

[266] Jing Wang and Xin Geng. 2019. Classification with Label Distribution Learning. In *Proceedings of the Twenty-Eighth International Joint Conference on Artificial Intelligence*. 3712–3718.

[267] Peng Wang, Meng Yan, Xiangping Zhan, Mei Tian, Yingdong Si, Yu Sun, Longzhen Jiao, and Xiaojie Wu. 2021. Predicting self-reported proactive personality classification with Weibo text and short answer text. *IEEE Access* 9 (2021), 77203–77211.

[268] Weichen Wang, Gabriella M Harari, Rui Wang, Sandrine R Müller, Shayan Mirjafari, Kizito Masaba, and Andrew T Campbell. 2018. Sensing behavioral change over time: Using within-person variability features from mobile sensing to predict personality traits. *Proceedings of the ACM on Interactive, Mobile, Wearable and Ubiquitous Technologies* 2, 3 (2018), 1–21.

[269] Xintao Wang, Yunze Xiao, Jen-tse Huang, Siyu Yuan, Rui Xu, Haoran Guo, Quan Tu, Yaying Fei, Ziang Leng, Wei Wang, Jiangjie Chen, Cheng Li, and Yanghua Xiao. 2024. InCharacter: Evaluating Personality Fidelity in Role-Playing Agents through Psychological Interviews. In *Proceedings of the 62nd Annual Meeting of the Association for Computational Linguistics (Volume 1: Long Papers)*, Lun-Wei Ku, Andre Martins, and Vivek Srikumar

(Eds.). Association for Computational Linguistics, Bangkok, Thailand, 1840–1873.

[270] Yilei Wang, Jiabao Zhao, Deniz S. Ones, Liang He, and Xin Xu. 2025. Evaluating the ability of large language models to emulate personality. *Scientific Reports* 15, 1 (2025), 519.

[271] Zhiyuan Wen, Jiannong Cao, Yu Yang, Haoli Wang, Ruosong Yang, and Shuaiqi Liu. 2023. DesPrompt: Personality-descriptive prompt tuning for few-shot personality recognition. *Information Processing & Management* 60, 5 (2023), 103422.

[272] David William and Derwin Suhartono. 2021. Text-based depression detection on social media posts: A systematic literature review. *Procedia Computer Science* 179 (2021), 582–589.

[273] Michael Wilson. 1988. MRC psycholinguistic database: Machine-usable dictionary, version 2.00. *Behavior research methods, instruments, & computers* 20, 1 (1988), 6–10.

[274] Bernhard Wolf. 2005. Brunswik's original lens model. *University of Landau, Germany* 9 (2005), 1–9.

[275] William R Wright and David N Chin. 2014. Personality profiling from text: introducing part-of-speech N-grams. In *User Modeling, Adaptation, and Personalization: 22nd International Conference, UMAP 2014, Aalborg, Denmark, July 7-11, 2014. Proceedings 22*. Springer, 243–253.

[276] Cao Xiao, David Mandell Freeman, and Theodore Hwa. 2015. Detecting clusters of fake accounts in online social networks. In *Proceedings of the 8th ACM Workshop on Artificial Intelligence and Security*. 91–101.

[277] Xiaoyu Xiong, Maurizio Filippone, and Alessandro Vinciarelli. 2016. Looking good with Flickr faves: Gaussian processes for finding difference makers in personality impressions. In *Proceedings of the 24th ACM international conference on Multimedia*. 412–415.

[278] Jia Xu, Weijian Tian, Guoyun Lv, Shiya Liu, and Yangyu Fan. 2021. Prediction of the big five personality traits using static facial images of college students with different academic backgrounds. *Ieee Access* 9 (2021), 76822–76832.

[279] Di Xue, Zheng Hong, Shize Guo, Liang Gao, Lifa Wu, Jinghua Zheng, and Nan Zhao. 2017. Personality recognition on social media with label distribution learning. *IEEE Access* 5 (2017), 13478–13488.

[280] Di Xue, Lifa Wu, Zheng Hong, Shize Guo, Liang Gao, Zhiyong Wu, Xiaofeng Zhong, and Jianshan Sun. 2018. Deep learning-based personality recognition from text posts of online social networks. *Applied Intelligence* 48 (2018), 4232–4246.

[281] Xia Xue, Jun Feng, and Xia Sun. 2021. Semantic-enhanced sequential modeling for personality trait recognition from texts. *Applied Intelligence* (2021), 1–13.

[282] CY Yaakub, N Sulaiman, and CW Kim. 2010. A study on personality identification using game based theory. In *2010 2nd International Conference on Computer Technology and Development*. IEEE, 732–734.

[283] Kosuke Yamada, Ryohei Sasano, and Koichi Takeda. 2019. Incorporating textual information on user behavior for personality prediction. In *Proceedings of the 57th annual meeting of the association for computational linguistics: student research workshop*. 177–182.

[284] Hsin-Chang Yang and Zi-Rui Huang. 2019. Mining personality traits from social messages for game recommender systems. *Knowledge-Based Systems* 165 (2019), 157–168.

[285] Kai Yang, Hui Yuan, and Raymond YK Lau. 2022. PsyCredit: an interpretable deep learning-based credit assessment approach facilitated by psychometric natural language processing. *Expert Systems with Applications* 198 (2022), 116847.

[286] Tal Yarkoni. 2010. Personality in 100,000 words: A large-scale analysis of personality and word use among bloggers. *Journal of research in personality* 44, 3 (2010), 363–373.

[287] Wen Yeye, Chen Deyuan, Li Baobin, Wang Xiaoyang, Liu Xiaoqian, and Zhu Tingshao. 2020. Predicting personality based on self-introduction video. *IFAC-PapersOnLine* 53, 5 (2020), 452–457.

[288] Wu Youyou, Michal Kosinski, and David Stillwell. 2015. Computer-based personality judgments are more accurate than those made by humans. *PNAS* 112, 4 (2015), 1036–1040.

[289] Jianguo Yu and Konstantin Markov. 2017. Deep learning based personality recognition from facebook status updates. In *2017 IEEE 8th international conference on awareness science and technology (iCAST)*. IEEE, 383–387.

[290] Brahim Zarouali, Tom Dobber, Guy De Pauw, and Claes de Vreese. 2022. Using a personality-profiling algorithm to investigate political microtargeting: assessing the persuasion effects of personality-tailored ads on social media. *Communication Research* 49, 8 (2022), 1066–1091.

[291] Gloria Zen, Bruno Lepri, Elisa Ricci, and Oswald Lanz. 2010. Space speaks: towards socially and personality aware visual surveillance. In *Proceedings of the 1st ACM international workshop on Multimodal pervasive video analysis*. 37–42.

[292] Shanyang Zhao, Sherri Grasmuck, and Jason Martin. 2008. Identity construction on Facebook: Digital empowerment in anchored relationships. *Computers in human behavior* 24, 5 (2008), 1816–1836.

[293] Xiaoming Zhao, Zhiwei Tang, and Shiqing Zhang. 2022. Deep personality trait recognition: a survey. *Frontiers in Psychology* 13 (2022), 839619.

[294] Lin Zhou and Wenjun An. 2022. Data classification of mental health and personality evaluation based on network deep learning. *Mobile Information Systems* 2022, 1 (2022), 9251598.

[295] Michelle X Zhou, Gloria Mark, Jingyi Li, and Huahai Yang. 2019. Trusting virtual agents: The effect of personality. *ACM Transactions on Interactive Intelligent Systems (TiiS)* 9, 2-3 (2019), 1–36.